\documentclass{article}

\usepackage{microtype}
\usepackage{graphicx}
\usepackage{subcaption}
\usepackage{booktabs} 

\usepackage{hyperref}

\usepackage[preprint]{icml2026}

\usepackage{amsmath}
\usepackage{amssymb}
\usepackage{mathtools}
\usepackage{amsthm}

\usepackage[capitalize,noabbrev]{cleveref}

\theoremstyle{plain}

\theoremstyle{definition}

\theoremstyle{remark}

\usepackage[textsize=tiny]{todonotes}

\icmltitlerunning{Frequency Bias and OOD Generalization in Neural Operators under a Variable-Coefficient Wave Equation}

\begin{document}

\twocolumn[
  \icmltitle{Frequency Bias and OOD Generalization in Neural Operators \\under a Variable-Coefficient Wave Equation}



  \icmlsetsymbol{equal}{*}

  \begin{icmlauthorlist}
    \icmlauthor{Runlong Xie}{yyy}
    \icmlauthor{An Luo}{sch}
  \end{icmlauthorlist}

  \icmlaffiliation{yyy}{Independent Researcher}
  \icmlaffiliation{sch}{School of Statistics, University of Minnesota, MN, USA}

  \icmlcorrespondingauthor{An Luo}{luo00318@umn.edu}

  \icmlkeywords{Neural Operator, Operator Learning, Wave Equations}

  \vskip 0.3in
]



\printAffiliationsAndNotice{}  

\begin{abstract}

Neural operators learn to map initial conditions to the terminal solution of partial differential equations (PDEs), providing a surrogate for the full operator mapping.
This enables rapid prediction across different input configurations. While recent neural operator architectures have demonstrated strong performance on diverse PDE tasks, their behavior under structured distribution shifts remains insufficiently understood. To investigate this, we study operator learning in a wave propagation setting governed by a one-dimensional variable-coefficient wave equation, using two representative architectures, the Fourier Neural Operator (FNO) and the Deep Operator Network (DeepONet). To examine their generalization under distribution shifts, we consider structured out-of-distribution (OOD) settings that independently vary input frequency and coefficient smoothness. The results show that under smoothness shifts, both models maintain stable performance, with FNO achieving lower error. In contrast, under frequency shifts, FNO exhibits a sharp increase in error under unseen high-frequency inputs, whereas DeepONet shows milder degradation despite higher overall error. Our analysis reveals that these differences arise from how each architecture represents and responds to variations in frequency structure. Together, these findings highlight a fundamental gap between strong in-distribution performance and generalization under distribution shifts in operator learning, underscoring the role of architectural representation bias in developing more reliable neural operators for physics-based PDE simulations beyond the training distribution.

\end{abstract}

\section{Introduction}

Neural operators learn mappings between input functions and PDE solutions, providing a surrogate for repeated numerical simulation \citep{lu2021learning,li2021fourier,kovachki2023neural}. Once trained, these models can rapidly predict solutions under different input configurations, making them attractive for scientific simulation, surrogate modeling, and computational physics \citep{kissas2020machine,pathak2022fourcastnet}. Recent neural-operator architectures have demonstrated strong empirical performance across a variety of PDE tasks, including fluid dynamics, transport problems, and wave propagation \citep{lu2021learning,li2021fourier,kovachki2023neural}.

Among existing neural operator architectures, the Fourier Neural Operator (FNO) and Deep Operator Network (DeepONet) are two representative approaches with substantially different operator representations. FNO constructs global representations through spectral convolution in Fourier space, allowing the model to efficiently learn couplings among frequency components \citep{li2020neural,li2021fourier}. DeepONet instead adopts a branch--trunk decomposition that combines representations of input functions and spatial coordinates to approximate function-to-function mappings \citep{lu2021learning,wang2021learning}. These models therefore provide two distinct perspectives on how neural operators represent PDE solution structure. Recent work has further extended neural operators toward larger-scale and more flexible PDE solvers, including physics-informed operator learning, graph-based operators, transformer-based PDE models, and pretrained foundation-style operator architectures \citep{li2021physics,cao2021global,rahman2022unified,wu2024transolver,herde2024poseidon,alkin2024upt,luo2025transolverpp}.

Despite these advances, the behavior of neural operators under
structured distribution shifts remains insufficiently understood
\citep{goswami2022deep,raonic2023convolutional}.
Existing studies have primarily focused on improving predictive
accuracy, scalability, or geometric flexibility
\citep{kovachki2023neural,li2023geometryinformed},
while comparatively less attention has been paid to how different
operator representations respond to systematic changes in input
structure \citep{raonic2023convolutional,liu2023incontext}.
In particular, it remains unclear whether strong in-distribution performance necessarily leads to stable extrapolation when the input frequency content or medium structure differs from the training distribution \citep{liu2023incontext,xu2025frequencyprinciple}. This question is important in real-world scientific applications, where models are often applied to physical systems that differ systematically from the data used for training. Meanwhile, increasingly complex neural operator models introduce additional interacting factors~\citep{willard2022integrating,kovachki2023neural}, making it harder to isolate how the underlying operator representation itself influences generalization behavior. We therefore focus on FNO and DeepONet as two canonical and structurally transparent neural operators that provide a controlled setting for studying how different operator representations affect PDE solution behavior.

To investigate this problem, we study a terminal-state operator learning task governed by a one-dimensional variable-coefficient wave equation, which captures a basic form of wave propagation in media with spatially varying material properties~\citep{aki2002quantitative,leveque2007finite}. Given an initial displacement field and a spatially varying wave-speed coefficient, the objective is to predict the wave solution at a fixed future time. To examine model generalization under distribution shift, we construct structured out-of-distribution (OOD) settings that independently vary input frequency and coefficient smoothness. Rather than focusing only on predictive accuracy, our goal is to understand why different neural-operator architectures exhibit distinct degradation behavior under structured changes in the input distribution \citep{brandstetter2022message,li2022dissipative}. We therefore combine standard error metrics with spectral analysis and structured OOD evaluation to study how operator representations interact with frequency structure.

The contributions of this work are threefold. First, we formulate a terminal-state operator learning task based on a conservative variable-coefficient wave equation and construct both in-distribution (ID) and structured OOD evaluation settings. Second, we provide a controlled empirical comparison between FNO and DeepONet under a unified experimental framework, enabling direct assessment of architectural effects. Third, we combine spectral-error analysis with structured OOD testing to study degradation behavior across frequency regimes, providing empirical insight into how neural-operator architectures generalize in physically structured PDE settings.

\subsection{Related Work}

\textbf{Wave Equations and Wave Propagation Modeling.}
Wave equations are fundamental models for describing the propagation of energy and signals in space and time, with broad applications in acoustics, electromagnetics, structural vibration, and seismic imaging \citep{morse1986theoretical,aki2002quantitative,graff2012wave}. In many practical systems, the propagation speed varies spatially rather than remaining constant. For instance, seismic waves traveling through layered media depend strongly on local material properties \citep{virieux1986p,tarantola1984inverse}. Such phenomena are commonly modeled using variable-coefficient wave equations, where spatially varying coefficients govern reflection, scattering, and local propagation behavior \citep{aki2002quantitative, leveque2007finite}. Many applications involving wave-type PDEs, including seismic inversion, uncertainty quantification, and design optimization, require repeated simulations under varying initial conditions or material parameters \citep{virieux2009overview,mosser2020stochastic}. Traditional numerical methods, such as finite-difference and finite-element solvers, typically recompute the solution for each new configuration, leading to rapidly increasing computational cost as physical complexity and parameter dimensionality grow \citep{leveque2002finite,strikwerda2004finite}. This computational burden motivates the development of operator learning methods that directly approximate mappings between input functions and solution fields.

\textbf{Neural Operator Architectures.}
Among existing neural operator architectures, FNO~\citep{li2021fourier} and DeepONet~\citep{lu2021learning} are two representative architectures with substantially different inductive structures. Building on these canonical models, several extensions have been proposed to improve physical consistency, geometric flexibility, and scalability. Physics-Informed Neural Operator (PINO) incorporates PDE residual constraints into the training process \citep{li2021physics}. Graph Neural Operator extends operator learning to irregular geometries using graph-based representations \citep{rahman2022unified}, while Galerkin Transformer employs attention mechanisms to improve long-range interaction modeling \citep{cao2021global}. More recent work has further explored large-scale PDE foundation models, including Transolver, POSEIDON, and UPT \citep{wu2024transolver,herde2024poseidon,alkin2024upt}. These developments have substantially improved model scalability and empirical performance. However, they also introduce additional interacting factors, including attention mechanisms, pretraining strategies, and latent-space compression, which makes it more difficult to isolate how the underlying operator representation itself influences generalization behavior. Therefore, in this paper we focus on the two representative architectures, FNO and DeepONet.

\textbf{Generalization and Distribution Shift in Scientific Machine Learning.}
Recent studies have increasingly emphasized that scientific machine learning models should be evaluated not only by predictive accuracy, but also by robustness under physically meaningful distribution shifts \citep{karniadakis2021physics,willard2022integrating}. Different neural operators may exhibit substantially different inductive biases, which can strongly affect stability and out-of-distribution (OOD) behavior \citep{kovachki2023neural,li2022dissipative}. In many scientific applications, reliable extrapolation beyond the training distribution is unavoidable, making structured OOD evaluation particularly important.

Existing work has primarily focused on benchmarking model performance on specific PDE tasks, while comparatively fewer studies systematically analyze how different neural operators respond to frequency variation and medium heterogeneity under a unified framework. In this work, we therefore focus on FNO and DeepONet as canonical and structurally transparent models, allowing us to study how different operator representations influence generalization behavior in variable-coefficient wave propagation.

\section{Problem Setup}

\subsection{Formulation of Operator Learning}

Many physical systems are modeled by PDEs, which describe how physical fields evolve over space and time. In scientific computing, these equations often need to be solved repeatedly under different initial conditions, boundary conditions, or material parameters. As system complexity increases, repeated numerical simulation can become computationally expensive.

Operator learning aims to reduce this cost by directly learning mappings from problem conditions to solution fields. Unlike standard supervised learning, which usually maps finite-dimensional vectors to vectors or scalars, operator learning focuses on mappings between functions. The objective is to learn how changes in input functions affect the structure of the output solution.

In this work, we study wave propagation in heterogeneous media using a one-dimensional conservative variable-coefficient wave equation. The spatial domain is \(x\in[0,1]\). The inputs are the initial displacement field \(u_0(x)\) and the spatially varying wave-speed function \(c(x)\). Given these inputs, the model predicts the wave field at a fixed terminal time \(T\), denoted by \(u(x,T)\). The corresponding operator mapping is given by
$\mathcal{G}:(u_0(x),c(x))\mapsto u(x,T).$

This setting provides a controlled environment for studying how neural-operator architectures behave under structured distribution shift. In particular, we focus on changes in input frequency and coefficient smoothness while minimizing other factors such as model scale or pretraining.

\subsection{The Variable-Coefficient Wave Equation}

The operator mapping considered in this paper is defined through the following one-dimensional conservative variable-coefficient wave equation:
$u_{tt}
=
\partial_x(c(x)^2u_x),
x\in[0,1], t\in[0,T].$ Here, \(u(x,t)\) denotes the wave displacement field, \(u_x=\partial u/\partial x\) denotes the spatial derivative of \(u\) and \(c(x)\) controls the local propagation speed. Unlike constant-coefficient wave equations, the propagation speed in the present system varies across space. This allows the equation to model heterogeneous media, where local material variation influences wave propagation, oscillation patterns, and energy transport.

We adopt the conservative form instead of the simplified equation \(u_{tt}=c(x)^2u_{xx}\). The conservative formulation better represents local interactions during propagation when the medium changes spatially.

Homogeneous Dirichlet boundary conditions are imposed as
$u(0,t)=u(1,t)=0,$
which correspond to fixed boundaries. The initial velocity is set to zero, i.e., 
$u_t(x,0)=0.$ This problem also exhibits strong frequency structure. Different frequency components in the input wave field interact with the spatially varying coefficient field during propagation, leading to complex wave behavior. This makes the problem a suitable testbed for studying frequency bias and OOD generalization in neural operators.

\section{Methods}

This section describes the neural architectures used in this study. Our focus is on how different model designs approximate mappings between functions and how their structural properties influence generalization. Additional implementation details, training procedures, and hyperparameter configurations are provided in the experimental settings section and Appendix~B.

\subsection{Fourier Neural Operator (FNO)}

FNO \citep{li2021fourier} approximates the operator using spectral convolution in Fourier space, enabling global interactions across the spatial domain. The model first lifts the input functions into a higher-dimensional latent representation through \(v_0(x)=P(u_0(x),c(x))\), where \(P\) denotes a learned lifting operator. The latent representation is then updated through multiple Fourier layers of the form
$v_{l+1}(x)=\sigma\!\left(Wv_l(x)+\mathcal{F}^{-1}\!\big(R(k)\cdot \mathcal{F}(v_l)(k)\big)\right),$
where \(v_l(x)\) denotes the latent feature representation at layer \(l\), \(\mathcal{F}\) and \(\mathcal{F}^{-1}\) denote the Fourier transform and inverse Fourier transform, \(k\) denotes the Fourier mode index, \(R(k)\) contains learnable spectral weights, and \(W\) is a pointwise linear transformation.

To reduce computational cost and impose an inductive bias toward smooth structures, FNO retains only a fixed number of low-frequency modes during spectral convolution. This truncation reflects the assumption that dominant solution structures are primarily encoded in low-frequency components while also improving computational efficiency. The final prediction is obtained through an output projection \(\hat{u}(x,T)=Q(v_L(x))\), where \(Q\) denotes a learned projection back to physical space.

\subsection{Deep Operator Network (DeepONet)}

DeepONet \citep{lu2021learning} constructs mappings using a branch--trunk decomposition, where one network encodes the input functions and the other encodes spatial coordinates. The branch network produces coefficients $b(u_b)$ from the input functions, while the trunk network generates basis functions $t(x)$ conditioned on spatial location. The output is computed as an inner product $\hat{u}(x,T)=\sum_{k=1}^K b_k(u_b)t_k(x)$, which can be interpreted as a learned expansion over basis functions. This formulation allows the model to represent complex solution structures in a coordinate-dependent manner. Unlike FNO, DeepONet does not rely on explicit frequency representations or fixed spectral truncation.

\subsection{Architectural Comparison Framework}

FNO and DeepONet embody fundamentally different inductive biases in how they represent operators. FNO models interactions between frequency components, while DeepONet constructs solutions through coordinate-conditioned basis expansions. These differences suggest that the two models may exhibit distinct behaviors under structured distribution shifts.

In this work, we compare these architectures in a controlled setting to isolate the effect of architectural design on generalization. Rather than incorporating more recent models that introduce additional factors such as large-scale pretraining, attention mechanisms, or multi-resolution designs, we deliberately focus on these canonical architectures. This allows observed differences in performance to be more directly attributed to underlying representation mechanisms. Both models are trained under identical data distributions and optimization settings to ensure a fair comparison.

\section{Experimental Setting}

\begin{figure*}[t]
    \centering
    \includegraphics[width=0.75\textwidth]
    {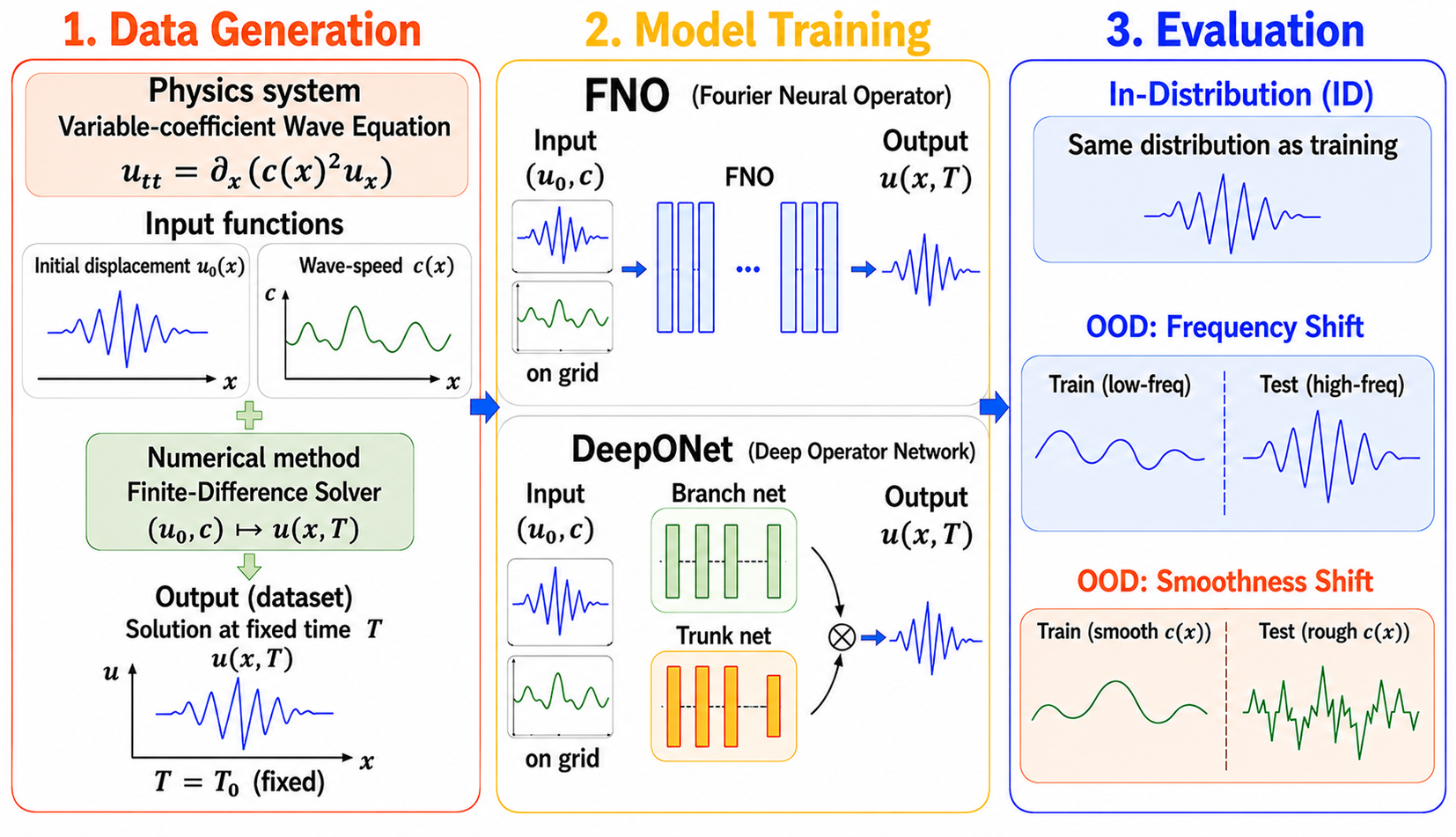}
    \caption{
    An overview of our experimental pipeline for operator learning under the variable-coefficient wave equation. 
    Input functions consist of an initial displacement field \(u_0(x)\) and a spatially varying wave-speed coefficient \(c(x)\). 
    A finite-difference solver generates the terminal solution \(u(x,T)\) at a fixed time horizon, forming supervised training data. 
    FNO and DeepONet are then trained to approximate the mapping from \((u_0,c)\) to \(u(x,T)\). 
    Evaluation is performed on in-distribution samples and two structured OOD settings: a frequency shift in the initial condition and a smoothness shift in the coefficient field.
    }
    \label{fig:experimental_pipeline}
\end{figure*}

This section describes the data generation process, dataset design, training procedures, and evaluation metrics used in our experiments. An overview of the experimental pipeline is presented in Figure~\ref{fig:experimental_pipeline}. The goal is to ensure a controlled and reproducible comparison between different neural operator architectures.

\subsection{Data Generation Pipeline}
\label{sec:data_generation}

The data are generated using a conservative finite-difference solver for the one-dimensional variable-coefficient wave equation. Each data sample is represented as $(u_0(x), c(x), u(x,T))$, where $u_0(x)$ is the initial displacement, $c(x)$ is the spatially varying wave-speed field, and $u(x,T)$ is the solution at a fixed terminal time $T$.

The numerical solver used for data generation follows a second-order finite-difference scheme in both space and time, meaning that the discretization error decreases quadratically as the grid resolution increases \citep{leveque2007finite,strikwerda2004finite}. The solver preserves the conservative structure of the PDE through a flux-based discretization. Numerical stability is maintained by satisfying the Courant--Friedrichs--Lewy (CFL) condition \citep{courant1967partial,leveque2002finite}, and harmonic averaging \citep{leveque2007finite,leveque2002finite} is used at grid interfaces to improve the treatment of spatially varying coefficients.

Initial displacement fields \(u_0(x)\) are generated using random Fourier combinations, i.e., random sums of sine functions with different amplitudes and frequencies. Training and ID samples use lower-frequency modes, while OOD-frequency samples additionally include higher-frequency components not observed during training. This ensures diversity in oscillatory structure while maintaining controlled frequency variation. The coefficient fields \(c(x)\) are constructed through random frequency superposition, meaning that multiple sinusoidal components with different spatial frequencies are combined to generate heterogeneous media. The resulting coefficient fields are grouped into three regimes: smooth, medium, and rough, corresponding to increasing levels of spatial variability. This setup allows controlled manipulation of input structure while maintaining physically meaningful wave dynamics.

\subsection{Dataset Partition with OOD Design}

The dataset is divided into training, ID testing, and OOD testing subsets. The ID set follows the same sampling rules as the training data, while the OOD sets are constructed by systematically modifying specific input characteristics.

Two types of OOD scenarios are considered. First, frequency-based shifts are introduced by generating initial displacement fields with higher-frequency components than those seen during training. Second, smoothness-based shifts are created by altering the spatial variability of the coefficient field \(c(x)\), as described in Section~\ref{sec:data_generation}. This setting tests model robustness under changes in the heterogeneity of the propagation medium.

These OOD settings are designed to reflect physically meaningful variations. Frequency shifts introduce finer-scale oscillatory structures into the initial wave field, corresponding to wave phenomena with shorter spatial wavelengths \citep{graff2012wave,morse1986theoretical}. Smoothness shifts instead modify the spatial variability of the coefficient field, reflecting changes in material heterogeneity and propagation properties across the domain \citep{aki2002quantitative,virieux2009overview}. Detailed sampling rules, parameter ranges, and dataset partitioning strategies are provided in Appendix~B.

\subsection{Model Setup and Training Procedure}

The two models considered in this paper, FNO and DeepONet, are trained under identical conditions to ensure a fair comparison. Both models are optimized using the relative \(L_2\) loss, defined as \(\mathcal{L}=\|u_{\mathrm{pred}}-u_{\mathrm{true}}\|_2/\|u_{\mathrm{true}}\|_2\), where \(u_{\mathrm{pred}}\) and \(u_{\mathrm{true}}\) denote the predicted and reference solutions, respectively. Optimization is performed using the Adam optimizer with a fixed learning rate. The same dataset splits, batch sizes, and training durations are used for both models, and all experiments are conducted with fixed random seeds to ensure reproducibility.

For FNO, key hyperparameters include the number of retained Fourier modes, hidden channel width, number of Fourier layers, and padding size used to mitigate boundary effects. For DeepONet, hyperparameters include the hidden dimensions of the branch and trunk networks, the size of the latent feature space, and network depth. Detailed hyperparameter settings and training configurations are provided in Appendix~B.

\subsection{Evaluation Metrics}

Model performance is primarily evaluated using the relative \(L_2\) error, defined as \(\|u_{\mathrm{pred}}-u_{\mathrm{true}}\|_2/\|u_{\mathrm{true}}\|_2\), which measures prediction error relative to the magnitude of the reference solution field.

To assess whether predicted solutions preserve physically meaningful spatial structure, we additionally compute an energy-like diagnostic quantity defined as \(E(u)=\int c(x)^2 (u_x)^2 dx\). This quantity is related to the spatial energy distribution commonly used in wave propagation and hyperbolic PDE analysis \citep{leveque2002finite,strauss2007partial}, where larger spatial gradients correspond to stronger local wave activity. Rather than serving as a strict conserved energy, it is used here as an auxiliary diagnostic for comparing the spatial structure of predicted and reference solutions.

We further perform spectral-error analysis by examining the Fourier spectrum of the prediction error \(e(x)=u_{\mathrm{pred}}(x)-u_{\mathrm{true}}(x)\). This analysis characterizes how prediction error is distributed across different frequency components and is particularly useful for studying degradation behavior under frequency-based distribution shifts.

All evaluation procedures are applied consistently across ID and OOD datasets. Additional experimental results and analysis details are provided in Appendix~C.

\section{Results and Analysis}

\begin{figure}[t]
    \centering
    \includegraphics[width=0.9\linewidth]{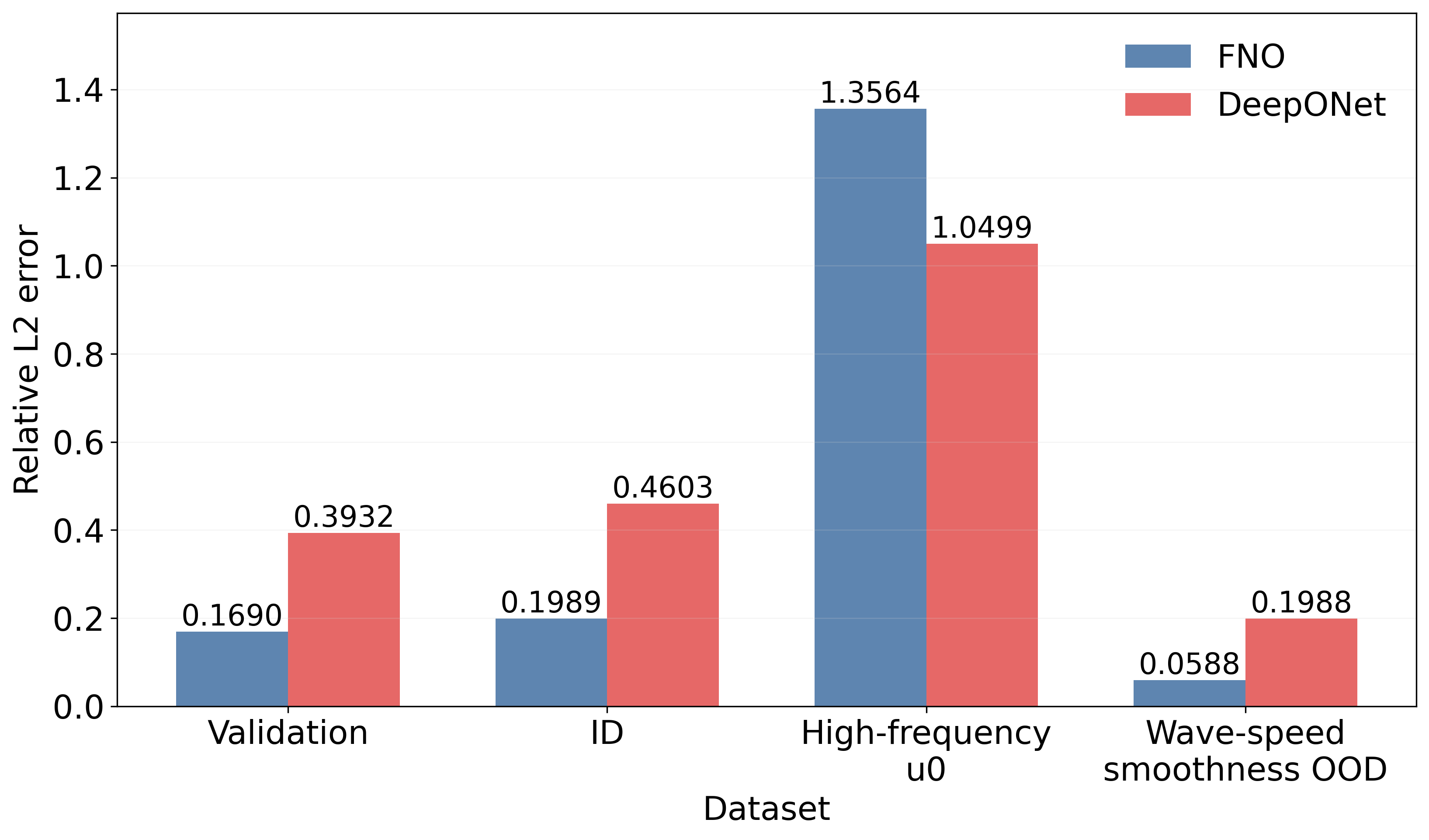}
    \caption{Comparison of relative \(L_2\) errors for FNO and DeepONet across Validation, in-distribution (ID), high-frequency OOD, and wave-speed smoothness OOD settings. FNO achieves lower error on Validation and ID datasets, but exhibits substantially larger degradation under high-frequency OOD, whereas DeepONet shows comparatively smoother degradation behavior.}
    \label{fig:main_results_bars}
\end{figure}

\begin{figure}[t]
    \centering
    \includegraphics[width=0.9\linewidth]{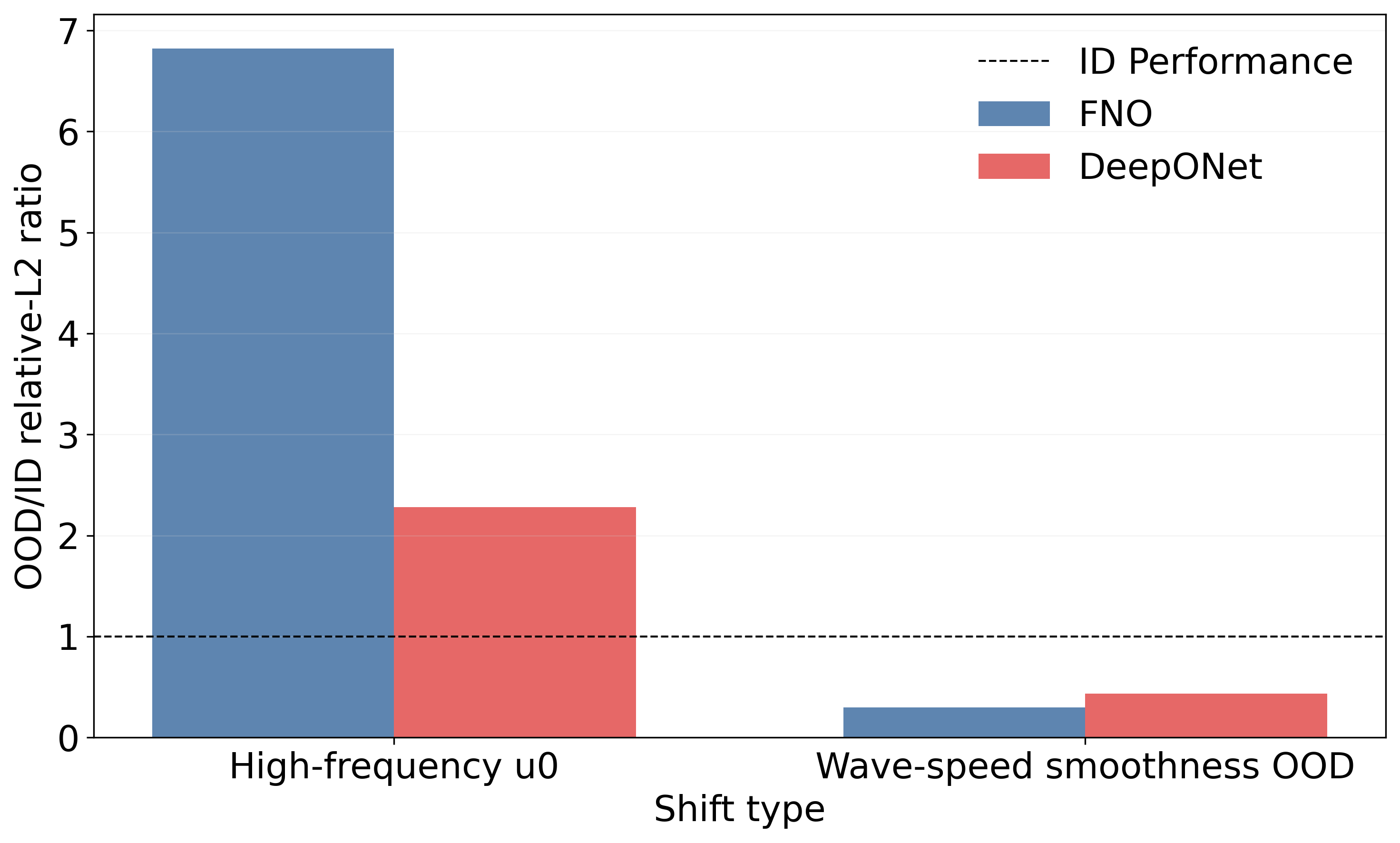}
    \caption{OOD-frequency degradation curves for FNO and DeepONet across all test samples. X-axis shows mode index $k$, Y-axis shows mean squared modal error. Left: ID spectral error; Right: OOD-frequency spectral error. FNO errors accumulate in multiple mid-to-high frequency modes under OOD-frequency conditions, while DeepONet exhibits smoother error growth. This figure provides a detailed view of how prediction errors are distributed across Fourier modes under distribution shift.}
    \label{fig:ood_degradation}
\end{figure}

We evaluate model performance under both ID and structured OOD settings, focusing on frequency and coefficient smoothness shifts. Our results reveal clear differences in how FNO and DeepONet generalize under these conditions. In particular, FNO achieves strong accuracy within the training distribution but degrades sharply under frequency shifts, whereas DeepONet exhibits more gradual performance changes. We further analyze these behaviors through spectral and spatial error decomposition to understand the underlying mechanisms.

\subsection{FNO Degrades Sharply under OOD-Frequency}

FNO shows a clear degradation pattern when the test inputs contain higher-frequency initial conditions than those observed during training. As shown in Figures~\ref{fig:main_results_bars} and~\ref{fig:ood_degradation}, its relative $L_2$ error increases substantially under the OOD-frequency setting compared with Validation and ID evaluation. This indicates that the strong in-distribution performance of FNO does not directly translate to stable extrapolation across input frequency regimes.

\begin{figure*}[t]
    \centering
    \includegraphics[width=\linewidth]{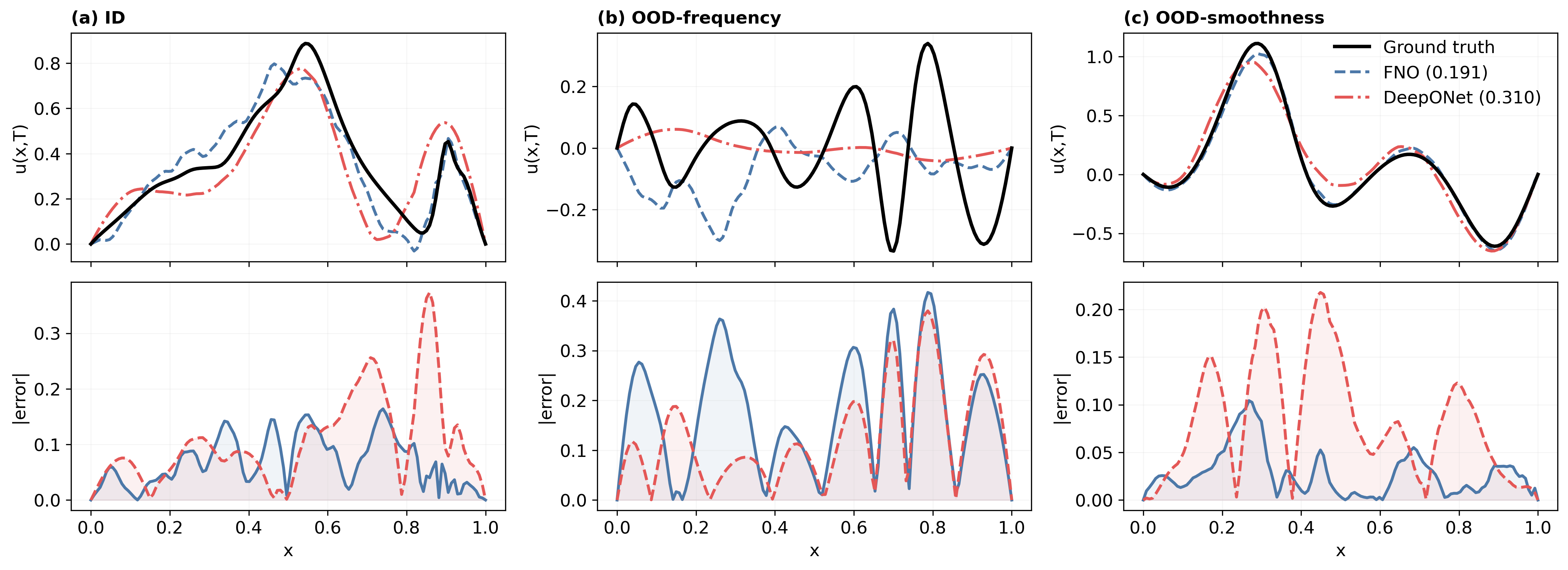}
    \caption{Representative prediction examples comparing FNO and DeepONet under different OOD conditions. Top row shows predicted terminal waveforms versus ground truth for ID, OOD-frequency, and OOD-smoothness inputs. Bottom row shows the corresponding pointwise prediction errors. The figure illustrates that FNO suffers larger local distortions and phase errors under high-frequency OOD, whereas DeepONet maintains smoother waveform structure, highlighting differences in error patterns between the two architectures.}
    \label{fig:representative_cases}
\end{figure*}

The representative cases in Figure~\ref{fig:representative_cases} provide a more detailed view of this degradation. Under ID and OOD-smoothness settings, FNO largely preserves the waveform shape and peak locations. In contrast, under OOD-frequency inputs, its predictions exhibit more pronounced phase shifts and oscillatory distortions. The errors are not uniformly distributed across the spatial domain, but are concentrated in regions where the target solution contains rapid oscillations or strong local curvature. This suggests that the failure mode is closely related to unresolved or poorly extrapolated high-frequency structures rather than random prediction noise.

\begin{figure}[t]
    \centering
    \includegraphics[width=\linewidth]{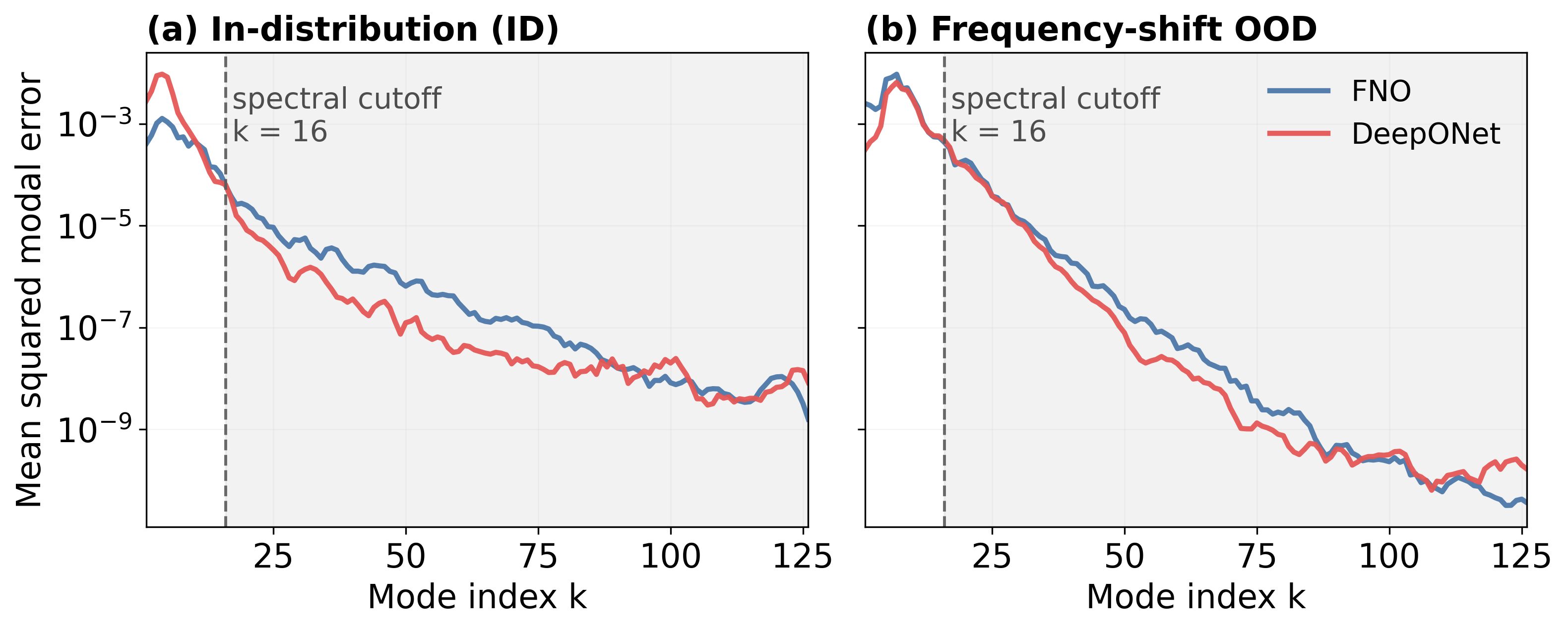}
    \caption{Spectral error analysis showing mean squared error per Fourier mode for FNO and DeepONet. Left: ID condition; Right: OOD-frequency condition. The figure shows that under ID conditions, errors decrease with increasing frequency and remain low for both models. Under OOD-frequency, FNO exhibits elevated error across multiple mid-to-high frequency modes, whereas DeepONet shows a more uniform and gradual error distribution. This illustrates how OOD-frequency inputs trigger multi-mode spectral errors in FNO.}
    \label{fig:spectral_error_distribution}
\end{figure}

The spectral-error analysis in Figure~\ref{fig:spectral_error_distribution} further supports the interpretation that FNO degradation under OOD-frequency is associated with instability in its learned spectral representations rather than failure at only the highest-frequency modes. Under OOD-frequency conditions, FNO error increases across a range of mid-to-high Fourier modes instead of appearing only at the largest modes. This pattern suggests that the degradation is not simply a pointwise loss of the highest-frequency components. Rather, unseen high-frequency inputs affect the broader spectral structure of the predicted solution, producing errors that spread across multiple frequency bands. This behavior is consistent with the architecture of FNO, where Fourier layers learn transformations among retained spectral components. When the input frequency distribution shifts beyond the range emphasized during training, the learned spectral relationships may no longer provide stable extrapolation.

\begin{figure}[t]
    \centering
    \includegraphics[width=0.9\linewidth]{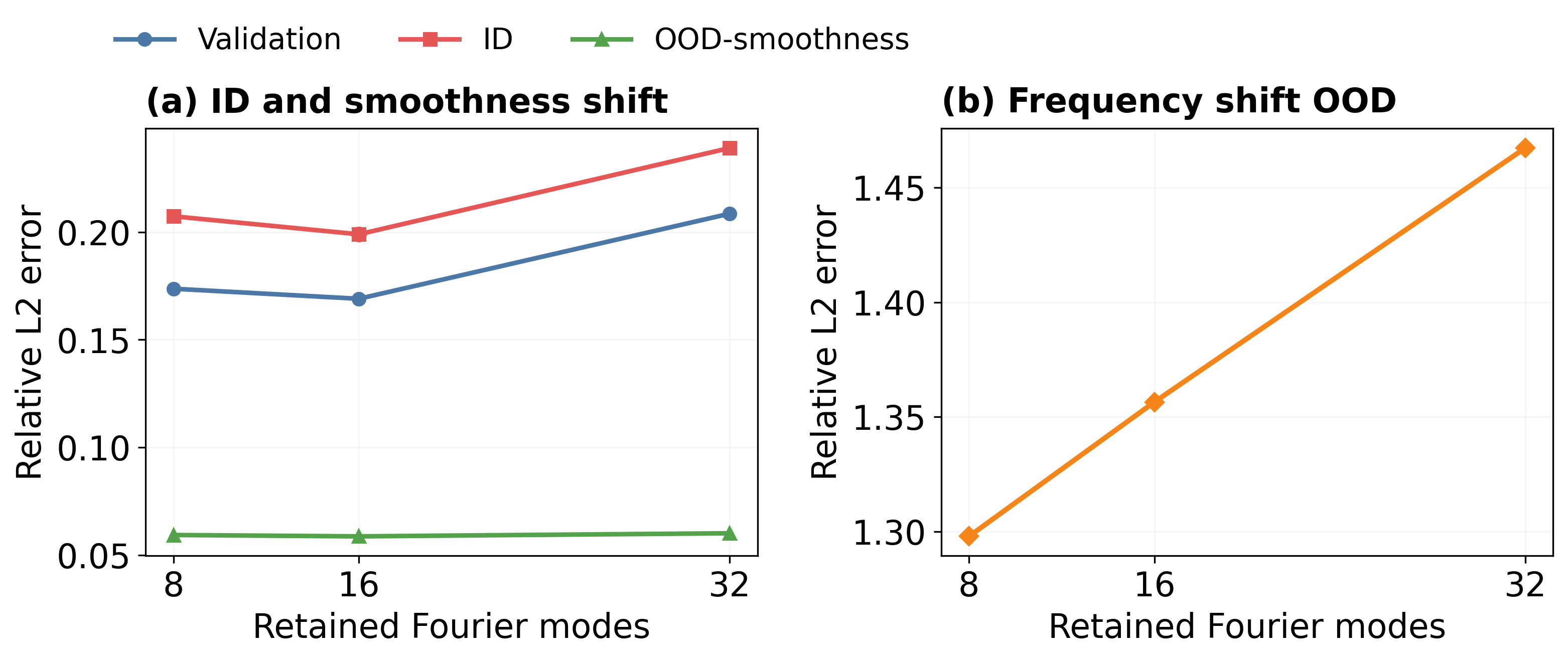}
    \caption{Ablation study on the number of retained Fourier modes in FNO. X-axis indicates retained mode count (8, 16, 32), and Y-axis shows relative $L_2$ error for ID (dashed line) and OOD-frequency (bars). While ID performance remains stable and optimal at intermediate modes, OOD-frequency error increases with more retained modes, demonstrating that simply increasing spectral capacity does not prevent OOD degradation. This figure supports that FNO's OOD-frequency limitations arise from architectural representation rather than insufficient model capacity.}
    \label{fig:modes_ablation}
\end{figure}

The retained-mode ablation in Figure~\ref{fig:modes_ablation} provides additional evidence that increasing spectral capacity alone does not resolve the observed OOD-frequency degradation. Increasing the number of retained Fourier modes from 8 to 16 and then to 32 does not reduce the OOD-frequency error; instead, the error increases, while ID performance remains stable and is best at an intermediate mode count. This result suggests that the OOD-frequency degradation is not merely caused by using too few Fourier modes. Additional spectral capacity alone does not guarantee better extrapolation when the training data do not constrain the corresponding high-frequency relationships. Thus, FNO's sharp degradation under OOD-frequency is better understood as a limitation of its spectral representation under distribution shift, rather than a simple capacity issue.

\subsection{DeepONet Exhibits Smoother Degradation under OOD-Frequency}

DeepONet exhibits a different degradation pattern under OOD-frequency conditions compared to its in-distribution behavior. As shown in Figures~\ref{fig:main_results_bars} and~\ref{fig:ood_degradation}, its prediction error increases when exposed to higher-frequency inputs, but the change is more gradual across samples, without the sharp performance drop observed for FNO. This suggests that its generalization behavior is less sensitive to abrupt shifts in input frequency.

The representative examples in Figure~\ref{fig:representative_cases} provide further insight into this behavior. Under OOD-frequency inputs, DeepONet is able to preserve the overall oscillatory structure of the waveform, including the approximate locations of peaks and troughs. Although local amplitude discrepancies and phase deviations are present, these errors do not accumulate into large localized distortions. Instead, the deviations appear more uniformly distributed across the spatial domain, indicating a smoother form of degradation.

This observation is consistent with the spectral-error analysis in Figure~\ref{fig:spectral_error_distribution}. Unlike FNO, where errors concentrate in specific mid-to-high Fourier modes, DeepONet shows a more evenly distributed error profile across frequencies. No single frequency band dominates the error, and the increase in error under OOD-frequency appears as a gradual shift rather than a concentrated spike.

A possible explanation for this behavior lies in the representation structure of DeepONet. By combining input-dependent coefficients with coordinate-dependent basis functions, the model effectively constructs solutions as a flexible expansion over learned basis functions. Since this representation does not rely on a fixed set of retained Fourier modes, it does not enforce a hard spectral cutoff, which may contribute to the smoother transition in error when encountering unseen frequency components. While this does not eliminate OOD degradation, it results in a more gradual and spatially distributed error pattern compared to FNO.

\subsection{Spatial Errors Correspond to Multi-Mode Spectral Deviation}

To better understand the degradation patterns observed under OOD-frequency conditions, we analyze prediction errors jointly in the spatial and spectral domains (Figures~\ref{fig:representative_cases} and~\ref{fig:spectral_error_distribution}). Across both models, errors tend to concentrate in spatial regions where the target solution exhibits strong oscillations or high local curvature. These regions are not isolated anomalies, but consistently correspond to structured deviations in the frequency domain.

For FNO, the OOD-frequency samples reveal a clear alignment between localized spatial errors and elevated error across multiple mid-to-high Fourier modes. Importantly, the error is not confined to the highest frequencies, but spreads across a band of modes. This suggests that the degradation cannot be attributed to a single missing frequency component. Instead, it reflects a broader disruption in how frequency components interact in the learned representation, leading to correlated errors across multiple modes.

DeepONet shows a different pattern. Although spatial errors also appear in high-oscillation regions, the corresponding spectral error is more evenly distributed, without a clear concentration in specific frequency bands. The absence of strong multi-mode accumulation indicates that the model does not exhibit the same structured failure across frequency components. Instead, its errors appear as smoother, distributed deviations.

Taken together, these observations highlight a consistent relationship between spatial and spectral error structure. Under OOD-frequency conditions, localized prediction errors are closely tied to how models represent and combine frequency information. The contrast between FNO and DeepONet suggests that differences in representation structure lead to qualitatively different error patterns, particularly in how errors propagate across frequency components.

\subsection{OOD Degradation is Closely Linked to Architectural Bias}

The observations from the previous subsections provide a consistent picture of how model behavior changes under OOD-frequency conditions. Both FNO and DeepONet achieve low and stable errors on Validation and ID datasets, indicating that the models successfully learn the mapping within the training distribution. However, their behaviors diverge significantly once the input frequency structure moves beyond this range.

For FNO, increasing the number of retained Fourier modes does not reduce OOD-frequency error (Figure~\ref{fig:modes_ablation}); instead, the error increases while ID performance remains stable. This suggests that simply expanding spectral capacity does not translate into improved generalization under frequency shift. Combined with the multi-mode spectral error patterns observed in Section~5.3, this behavior points to a limitation in how the learned spectral representation extrapolates beyond the training regime.

For DeepONet, although the overall error is higher, the degradation remains gradual and spatially distributed. The absence of concentrated spectral error accumulation suggests a different interaction between the model representation and input frequency structure.

One possible interpretation is that OOD-frequency degradation is influenced not only by the availability of training data, but also by how each architecture encodes and extrapolates frequency information. While extending the training distribution to include higher-frequency samples could improve performance within that range, the observed patterns indicate that model behavior under unseen frequency conditions is shaped by the underlying representation mechanism.

Overall, the combination of stable in-distribution performance, insensitivity of OOD behavior to increased spectral capacity, and structured spectral error patterns suggests that architectural bias plays a central role in determining generalization under frequency shift. These findings provide a unified perspective for understanding the different OOD behaviors of FNO and DeepONet.

\section{Conclusion}

This work studies terminal-state operator learning for a one-dimensional variable-coefficient wave equation, with a focus on how neural operator architectures influence generalization. We conduct a controlled comparison between FNO and DeepONet under structured in-distribution and out-of-distribution settings. The results show that FNO achieves higher accuracy within the training distribution and under smoothness shifts, while DeepONet exhibits more stable performance under frequency OOD conditions. Further analysis reveals that these differences are closely related to the underlying architectural biases, particularly in how each model represents and extrapolates frequency components.

These findings highlight that predictive accuracy alone is insufficient to characterize neural operator performance in physics-based settings. Structured OOD evaluation and spectral analysis provide deeper insight into how architectures for operator learning interact with input structure. In particular, the observed degradation patterns suggest that generalization is strongly influenced by how frequency information is encoded and coupled within the model.

Future work may explore architectures for operator learning that explicitly adapt to the spectral properties of input functions. For example, frequency-aware representations or learnable spectral support could allow models to dynamically adjust to unseen frequency regimes. Another promising direction is the integration of adaptive or hybrid representations that combine global spectral modeling with localized spatial features, potentially improving robustness while maintaining computational efficiency.

\section*{Impact Statement}

This work studies neural operator learning for variable-coefficient wave equations, with a focus on understanding generalization behavior under structured distribution shifts. Neural operators have the potential to accelerate repeated PDE evaluation in scientific computing, physics simulation, and engineering design, where fast approximation of solution operators may reduce computational cost.

At the same time, our results suggest that neural operator performance can depend strongly on the frequency structures represented during training. In particular, models that perform well within the training distribution may still degrade under unseen frequency conditions. This highlights the importance of evaluating robustness beyond standard in-distribution metrics before deploying neural operators in scientific or engineering settings.

We do not identify direct ethical concerns associated with this work. However, understanding the limitations and failure modes of learned PDE surrogates may contribute to safer and more reliable use of machine learning methods in physics-informed applications.


\nocite{langley00}

\bibliography{example_paper}

@inproceedings{langley00,
 author    = {P. Langley},
 title     = {Crafting Papers on Machine Learning},
 year      = {2000},
 pages     = {1207--1216},
 editor    = {Pat Langley},
 booktitle     = {Proceedings of the 17th International Conference
              on Machine Learning (ICML 2000)},
 address   = {Stanford, CA},
 publisher = {Morgan Kaufmann}
}

@book{leveque2002finite,
  title     = {Finite Volume Methods for Hyperbolic Problems},
  author    = {LeVeque, Randall J.},
  year      = {2002},
  publisher = {Cambridge University Press}
}

@book{strikwerda2004finite,
  title     = {Finite Difference Schemes and Partial Differential Equations},
  author    = {Strikwerda, John C.},
  edition   = {2},
  year      = {2004},
  publisher = {SIAM}
}

@book{strauss2007partial,
  title     = {Partial Differential Equations: An Introduction},
  author    = {Strauss, Walter A.},
  edition   = {2},
  year      = {2007},
  publisher = {Wiley}
}

@article{karniadakis2021physics,
  title   = {Physics-informed machine learning},
  author  = {Karniadakis, George Em and Kevrekidis, Ioannis G. and Lu, Lu and Perdikaris, Paris and Wang, Sifan and Yang, Liu},
  journal = {Nature Reviews Physics},
  volume  = {3},
  number  = {6},
  pages   = {422--440},
  year    = {2021}
}

@article{willard2022integrating,
  title   = {Integrating scientific knowledge with machine learning for engineering and environmental systems},
  author  = {Willard, Jared and Jia, Xiaowei and Xu, Shaoming and Steinbach, Michael and Kumar, Vipin},
  journal = {ACM Computing Surveys},
  volume  = {55},
  number  = {4},
  pages   = {1--37},
  year    = {2022}
}

@article{lu2021learning,
  title   = {Learning nonlinear operators via {DeepONet} based on the universal approximation theorem of operators},
  author  = {Lu, Lu and Jin, Pengzhan and Pang, Guofei and Zhang, Zhongqiang and Karniadakis, George Em},
  journal = {Nature Machine Intelligence},
  volume  = {3},
  number  = {3},
  pages   = {218--229},
  year    = {2021}
}

@inproceedings{li2021fourier,
  title     = {Fourier Neural Operator for Parametric Partial Differential Equations},
  author    = {Li, Zongyi and Kovachki, Nikola and Azizzadenesheli, Kamyar and Liu, Burigede and Bhattacharya, Kaushik and Stuart, Andrew and Anandkumar, Anima},
  booktitle = {International Conference on Learning Representations},
  year      = {2021}
}

@article{kovachki2023neural,
  title   = {Neural operator: Learning maps between function spaces with applications to {PDEs}},
  author  = {Kovachki, Nikola and Li, Zongyi and Liu, Burigede and Azizzadenesheli, Kamyar and Bhattacharya, Kaushik and Stuart, Andrew and Anandkumar, Anima},
  journal = {Journal of Machine Learning Research},
  volume  = {24},
  number  = {89},
  pages   = {1--97},
  year    = {2023}
}

@article{pathak2022fourcastnet,
  title   = {{FourCastNet}: A global data-driven high-resolution weather model using adaptive {Fourier} neural operators},
  author  = {Pathak, Jaideep and Subramanian, Shashank and Harrington, Peter and Raja, Sanjeev and Chattopadhyay, Ashesh and Mardani, Morteza and Kurth, Thorsten and Hall, David and Li, Zongyi and Azizzadenesheli, Kamyar and Hassanzadeh, Pedram and Kashinath, Karthik and Anandkumar, Anima},
  journal = {arXiv preprint arXiv:2202.11214},
  year    = {2022}
}

@article{kissas2020machine,
  title   = {Machine learning in cardiovascular flows modeling: Predicting arterial blood pressure from non-invasive 4D flow {MRI} data using physics-informed neural networks},
  author  = {Kissas, Georgios and Yang, Yaohua and Hwuang, En-Jui and Witschey, Walter R. and Detre, John A. and Perdikaris, Paris},
  journal = {Computer Methods in Applied Mechanics and Engineering},
  volume  = {358},
  pages   = {112623},
  year    = {2020}
}

@inproceedings{li2020neural,
  title     = {Neural Operator: Graph Kernel Network for Partial Differential Equations},
  author    = {Li, Zongyi and Kovachki, Nikola and Azizzadenesheli, Kamyar and Liu, Burigede and Bhattacharya, Kaushik and Stuart, Andrew and Anandkumar, Anima},
  booktitle = {ICLR Workshop on Integration of Deep Neural Models and Differential Equations},
  year      = {2020}
}

@article{wang2021learning,
  title   = {Learning the solution operator of parametric partial differential equations with physics-informed {DeepONets}},
  author  = {Wang, Sifan and Wang, Hanwen and Perdikaris, Paris},
  journal = {Science Advances},
  volume  = {7},
  number  = {40},
  pages   = {eabi8605},
  year    = {2021}
}

@article{li2021physics,
  title   = {Physics-informed neural operator for learning partial differential equations},
  author  = {Li, Zongyi and Zheng, Hongkai and Kovachki, Nikola and Jin, David and Chen, Haoxuan and Liu, Burigede and Azizzadenesheli, Kamyar and Anandkumar, Anima},
  journal = {arXiv preprint arXiv:2111.03794},
  year    = {2021}
}

@inproceedings{cao2021global,
  title     = {Choose a Transformer: Fourier or Galerkin},
  author    = {Cao, Shuhao},
  booktitle = {Advances in Neural Information Processing Systems},
  volume    = {34},
  pages     = {24924--24940},
  year      = {2021}
}

@article{rahman2022unified,
  title   = {{U-NO}: {U}-shaped Neural Operators},
  author  = {Rahman, Md Ashiqur and Ross, Zachary E. and Azizzadenesheli, Kamyar},
  journal = {Transactions on Machine Learning Research},
  year    = {2023}
}

@inproceedings{brandstetter2022message,
  title     = {Message Passing Neural {PDE} Solvers},
  author    = {Brandstetter, Johannes and Worrall, Daniel E. and Welling, Max},
  booktitle = {International Conference on Learning Representations},
  year      = {2022}
}

@inproceedings{li2022dissipative,
  title={Learning Dissipative Dynamics in Chaotic Systems},
  author={Li, Zongyi and Liu-Schiaffini, Miguel and Kovachki, Nikola and Liu, Burigede and Azizzadenesheli, Kamyar and Bhattacharya, Kaushik and Stuart, Andrew and Anandkumar, Anima},
  booktitle={Advances in Neural Information Processing Systems},
  volume={35},
  pages={16084--16097},
  year={2022}
}

@book{morse1986theoretical,
  title     = {Theoretical Acoustics},
  author    = {Morse, Philip M. and Ingard, K. Uno},
  year      = {1986},
  publisher = {Princeton University Press}
}

@book{graff2012wave,
  title     = {Wave Motion in Elastic Solids},
  author    = {Graff, Karl F.},
  year      = {2012},
  publisher = {Dover Publications}
}

@book{aki2002quantitative,
  title     = {Quantitative Seismology},
  author    = {Aki, Keiiti and Richards, Paul G.},
  edition   = {2},
  year      = {2002},
  publisher = {University Science Books}
}

@article{virieux2009overview,
  title   = {An overview of full-waveform inversion in exploration geophysics},
  author  = {Virieux, Jean and Operto, St{\'e}phane},
  journal = {Geophysics},
  volume  = {74},
  number  = {6},
  pages   = {WCC1--WCC26},
  year    = {2009}
}

@article{virieux1986p,
  title   = {{P-SV} wave propagation in heterogeneous media: Velocity-stress finite-difference method},
  author  = {Virieux, Jean},
  journal = {Geophysics},
  volume  = {51},
  number  = {4},
  pages   = {889--901},
  year    = {1986}
}

@article{tarantola1984inverse,
  title   = {Inversion of seismic reflection data in the acoustic approximation},
  author  = {Tarantola, Albert},
  journal = {Geophysics},
  volume  = {49},
  number  = {8},
  pages   = {1259--1266},
  year    = {1984}
}

@book{leveque2007finite,
  title     = {Finite Difference Methods for Ordinary and Partial Differential Equations: Steady-State and Time-Dependent Problems},
  author    = {LeVeque, Randall J.},
  year      = {2007},
  publisher = {SIAM}
}

@article{mosser2020stochastic,
  title   = {Stochastic seismic waveform inversion using generative adversarial networks as a geological prior},
  author  = {Mosser, Lukas and Dubrule, Olivier and Blunt, Martin J.},
  journal = {Mathematical Geosciences},
  volume  = {52},
  pages   = {53--79},
  year    = {2020}
}

@inproceedings{wu2024transolver,
  title     = {Transolver: A Fast Transformer Solver for PDEs on General Geometries},
  author    = {Wu, Haixu and Luo, Huakun and Wang, Haowen and Wang, Jianmin and Long, Mingsheng},
  booktitle = {Proceedings of the 41st International Conference on Machine Learning},
  year      = {2024}
}

@inproceedings{herde2024poseidon,
  title     = {{POSEIDON}: Efficient Foundation Models for {PDEs}},
  author    = {Herde, Maximilian and Raoni{\'c}, Bogdan and Rohner, Tobias and K{\"a}ppeli, Roger and Molinaro, Roberto and de B{\'e}zenac, Emmanuel and Mishra, Siddhartha},
  booktitle = {Advances in Neural Information Processing Systems},
  year      = {2024}
}

@inproceedings{alkin2024upt,
  title     = {Universal Physics Transformers: A Framework for Efficiently Scaling Neural Operators},
  author    = {Alkin, Benedikt and F{\"u}rst, Andreas and Schmid, Simon and Gruber, Lukas and Holzleitner, Markus and Brandstetter, Johannes},
  booktitle = {Advances in Neural Information Processing Systems},
  year      = {2024}
}

@misc{luo2025transolverpp,
  title        = {An Accurate Neural Solver for {PDEs} on Million-Scale Geometries},
  author       = {Luo, Hongkai and others},
  year         = {2025},
  note         = {OpenReview},
  url          = {https://openreview.net/forum?id=AM7iAh0krx}
}

@book{courant1967partial,
  title     = {Methods of Mathematical Physics, {Vol.}~2: Partial Differential Equations},
  author    = {Courant, Richard and Hilbert, David},
  year      = {1962},
  publisher = {Wiley-Interscience}
}

@article{goswami2022deep,
  author  = {Somdatta Goswami and Katiana Kontolati and
             Michael D. Shields and George Em Karniadakis},
  title   = {Deep transfer operator learning for partial differential
             equations under conditional shift},
  journal = {Nature Machine Intelligence},
  year    = {2022},
  doi     = {10.1038/s42256-022-00569-2}
}

@inproceedings{raonic2023convolutional,
  author    = {Bogdan Raoni{\'c} and Roberto Molinaro and
               Tim De~Ryck and Tobias Rohner and
               Francesca Bartolucci and Rima Alaifari and
               Siddhartha Mishra and Emmanuel de~B{\'e}zenac},
  title     = {Convolutional Neural Operators for Robust and Accurate
               Learning of {PDEs}},
  booktitle = {Advances in Neural Information Processing Systems},
  volume    = {36},
  year      = {2023},
  url       = {https://arxiv.org/abs/2302.01178}
}

@inproceedings{li2023geometryinformed,
  author    = {Zongyi Li and Nikola Kovachki and Chris Choy and
               Boyi Li and Jean Kautz and Anima Anandkumar},
  title     = {Geometry-Informed Neural Operator for Large-Scale
               {3D} {PDEs}},
  booktitle = {Advances in Neural Information Processing Systems},
  volume    = {36},
  year      = {2023},
  url       = {https://arxiv.org/abs/2309.00583}
}

@article{xu2025frequencyprinciple,
  author  = {Zhi-Qin John Xu and Yaoyu Zhang and Tao Luo},
  title   = {Overview Frequency Principle/Spectral Bias in Deep
             Learning},
  journal = {Communications on Applied Mathematics and Computation},
  year    = {2025},
  doi     = {10.48550/arXiv.2201.07395},
  url     = {https://arxiv.org/abs/2201.07395}
}

@inproceedings{liu2023incontext,
  author    = {Jerry Weihong Liu and N. Benjamin Erichson and
               Kush Bhatia and Michael W. Mahoney and
               Christopher R{\'e}},
  title     = {Does In-Context Operator Learning Generalize to
               Domain-Shifted Settings?},
  booktitle = {NeurIPS 2023 Workshop on The Symbiosis of Deep
               Learning and Differential Equations ({DLDE-III})},
  year      = {2023},
  url       = {https://openreview.net/forum?id=FsJ2FdQ19C}
}
\bibliographystyle{icml2026}

\newpage
\appendix
\onecolumn

\section{Numerical Solver Details}

This appendix provides additional details on the numerical reference solver used to generate the supervised learning data. Since the main focus of this work is neural-operator generalization rather than numerical discretization, implementation details of the solver are reported here for reproducibility.

All training and testing samples are generated using a unified finite-difference reference solver. The conservative variable-coefficient wave equation considered in this work is
\[
u_{tt}=\partial_x(c(x)^2u_x),
\]
defined on \(x\in[0,1]\), with homogeneous Dirichlet boundary conditions
\[
u(0,t)=u(1,t)=0.
\]
The initial conditions are
\[
u(x,0)=u_0(x),
\qquad
u_t(x,0)=0.
\]
Given an initial condition and a wave-speed function \(c(x)\), the reference solver computes the numerical solution at the fixed terminal time \(T\).

The spatial domain is discretized on a uniform grid. Let \(\Delta x\) denote the spatial step size, \(\Delta t\) the time step size, \(N_x\) the number of spatial grid points, and \(N_t\) the number of time steps.

To preserve the conservative structure of the PDE, the spatial discretization uses a numerical-flux finite-difference form. The spatial operator at grid point \(x_i\) is approximated by
\[
\partial_x(c(x)^2u_x)
\approx
\frac{
F_{i+\frac12}-F_{i-\frac12}
}{
\Delta x
},
\]
where the numerical flux is defined as
\[
F_{i+\frac12}
=
c_{i+\frac12}^2
\frac{u_{i+1}-u_i}{\Delta x}.
\]
Compared with directly discretizing \(c(x)^2u_{xx}\), this form more naturally follows the flux structure of the conservative PDE and is better suited for spatially varying coefficients.

Time integration is performed using a leapfrog scheme:
\[
u_i^{n+1}
=
2u_i^n-u_i^{n-1}
+
\Delta t^2
\left(
\partial_x(c(x)^2u_x)
\right)_i^n.
\]
This method is second-order accurate in time and is well suited for non-dissipative wave propagation.

Since the leapfrog method requires two initial time levels, the first step is initialized using a Taylor expansion:
\[
u_i^1
=
u_i^0
+
\frac{\Delta t^2}{2}
\left(
\partial_x(c(x)^2u_x)
\right)_i^0.
\]

The time step satisfies the CFL condition
\[
\max_x c(x)
\frac{\Delta t}{\Delta x}
\le C_{\mathrm{CFL}},
\]
which ensures that the numerical propagation speed remains within the stability range allowed by the discrete grid.

Before being used for data generation, the reference solver was verified in the constant-coefficient setting to confirm that its propagation behavior is consistent with the theoretical wave dynamics. The same solver is then used to generate all supervised learning samples.

\section{Dataset Sampling and OOD Construction}

This appendix provides additional details on dataset sampling and the construction of structured out-of-distribution splits.

Each sample consists of an input pair \((u_0(x),c(x))\) and the corresponding terminal solution \(u(x,T)\) generated by the reference solver.

Initial conditions are generated using random Fourier-mode combinations:
\[
u_0(x)
=
\sum_{k=1}^{K_u}
a_k\sin(k\pi x),
\]
where the random coefficients \(a_k\) control the contribution of each frequency component, and the frequency cutoff \(K_u\) controls the complexity of the initial condition.

During training, only low-frequency initial conditions are sampled:
\[
K_u \le K_{\mathrm{train}}.
\]
The OOD-frequency split is constructed by introducing higher-frequency initial conditions:
\[
K_u > K_{\mathrm{train}}.
\]
Thus, this split evaluates model behavior when the input contains oscillatory structures not observed during training.

The wave-speed function is generated through random frequency superposition:
\[
c(x)
=
1+
\sum_{k=1}^{K_c}
b_k\sin(2\pi kx+\phi_k),
\]
where \(b_k\) denotes the random amplitude, \(\phi_k\) denotes the random phase, and \(K_c\) controls the spatial smoothness of the coefficient field.

Depending on the range of \(K_c\), coefficient fields are categorized into smooth, medium, and rough regimes. The training and ID test sets are sampled from the same distribution, while the OOD-smoothness split changes the coefficient smoothness regime during testing.

Training, validation, ID test, and OOD test samples are generated using independent random seeds to avoid data leakage.

\section{Model and Training Hyperparameters}

This appendix reports the model and training configurations used in the main experiments. All models are trained using the Adam optimizer without learning-rate scheduling.

The supervised loss is the relative \(L_2\) loss:
\[
\mathcal{L}
=
\frac{
\|u_{\mathrm{pred}}-u_{\mathrm{true}}\|_2
}{
\|u_{\mathrm{true}}\|_2
}.
\]

\begin{table}[h]
\centering
\caption{Model and training hyperparameters used in the experiments.}
\label{tab:appendix_hyperparameters}
\begin{tabular}{lcc}
\toprule
Hyperparameter & FNO & DeepONet \\
\midrule
Learning rate & $1\times 10^{-3}$ & $1\times 10^{-3}$ \\
Batch size & 32 & 32 \\
Maximum epochs & 100 & 100 \\
Optimizer & Adam & Adam \\
Weight decay & 0 & 0 \\
Loss function & Relative $L_2$ & Relative $L_2$ \\
Learning-rate scheduler & None & None \\
Early-stopping patience & 10 & 10 \\
Random seed & 2026 & 2026 \\
Grid points & 128 & 128 \\
Retained Fourier modes & 16 & -- \\
Hidden width & 64 & -- \\
Fourier layers & 4 & -- \\
Padding size & 0 & -- \\
Activation & GELU & ReLU \\
Branch hidden dimension & -- & 128 \\
Trunk hidden dimension & -- & 128 \\
Latent dimension & -- & 128 \\
Network depth & -- & 3 \\
Output bias & -- & Enabled \\
Dirichlet output envelope & Enabled & Enabled \\
\bottomrule
\end{tabular}
\end{table}

Weight decay is reported as zero because the implementation uses PyTorch Adam without an explicit \texttt{weight\_decay} argument. The Dirichlet output envelope refers to a multiplicative boundary mask used to enforce zero-valued boundary behavior in model outputs.

\section{Additional Experimental Evidence}

This appendix provides supplementary evidence supporting the main observations reported in the paper. The additional figures and tables aim to show that the reported trends are consistent across multiple representative samples, evaluation settings, and frequency-domain analyses, rather than being based on a single selected example.

\subsection{Additional Frequency OOD Examples}

\begin{figure}[h]
    \centering
    \includegraphics[width=0.32\linewidth]{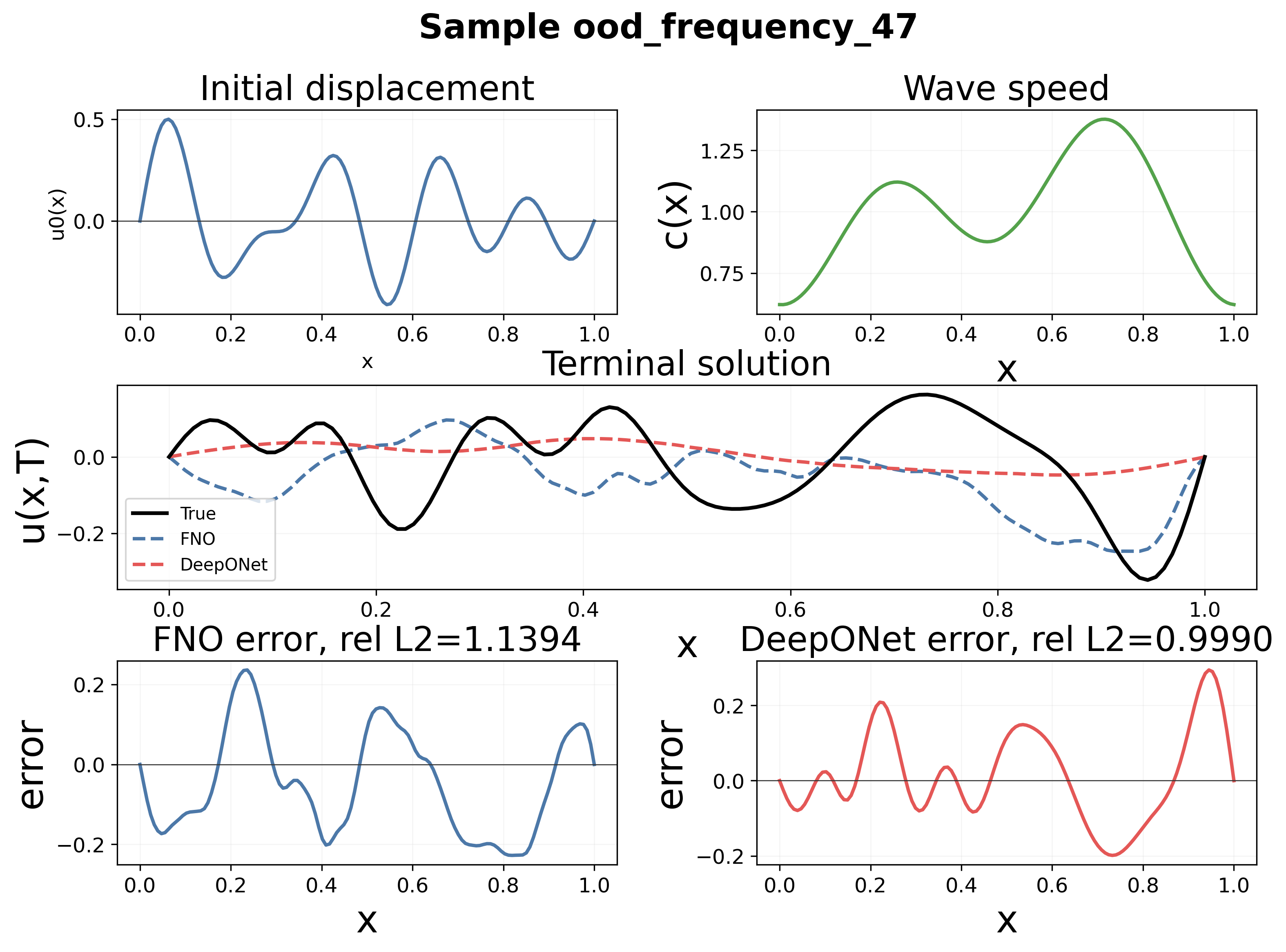}
    \includegraphics[width=0.32\linewidth]{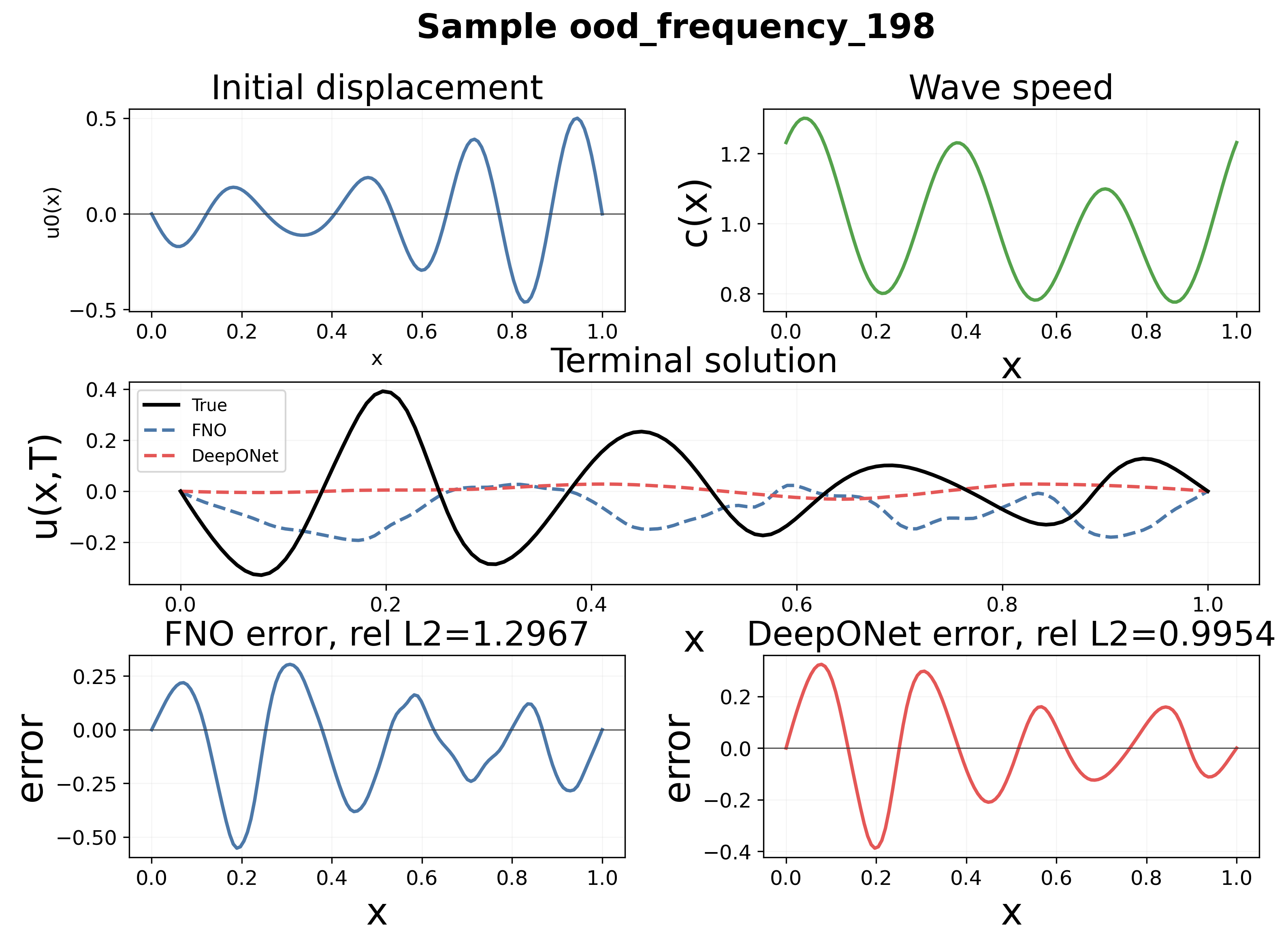}
    \includegraphics[width=0.32\linewidth]{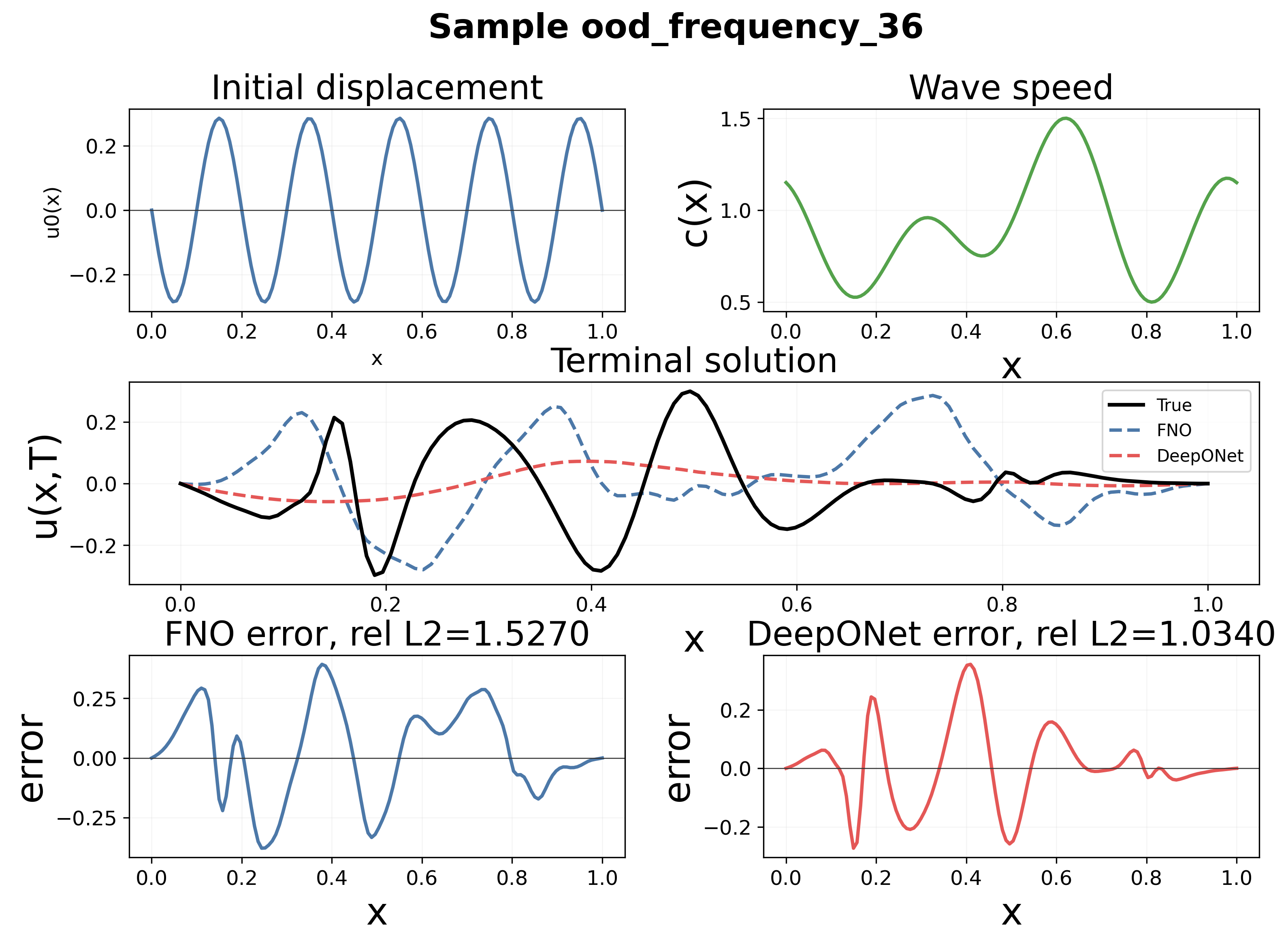}
    \caption{Additional OOD-frequency prediction examples illustrating representative failure patterns under high-frequency initial conditions. Each sample includes the initial condition, spatial coefficient field, terminal prediction, and residual distribution.}
    \label{fig:appendix_frequency_ood_examples}
\end{figure}

Figure~\ref{fig:appendix_frequency_ood_examples} shows additional high-frequency OOD samples. These examples use the same evaluation pipeline as the main results and provide a broader view of the frequency-shift behavior.

\subsection{Additional Smoothness OOD Examples}

\begin{figure}[h]
    \centering
    \includegraphics[width=0.32\linewidth]{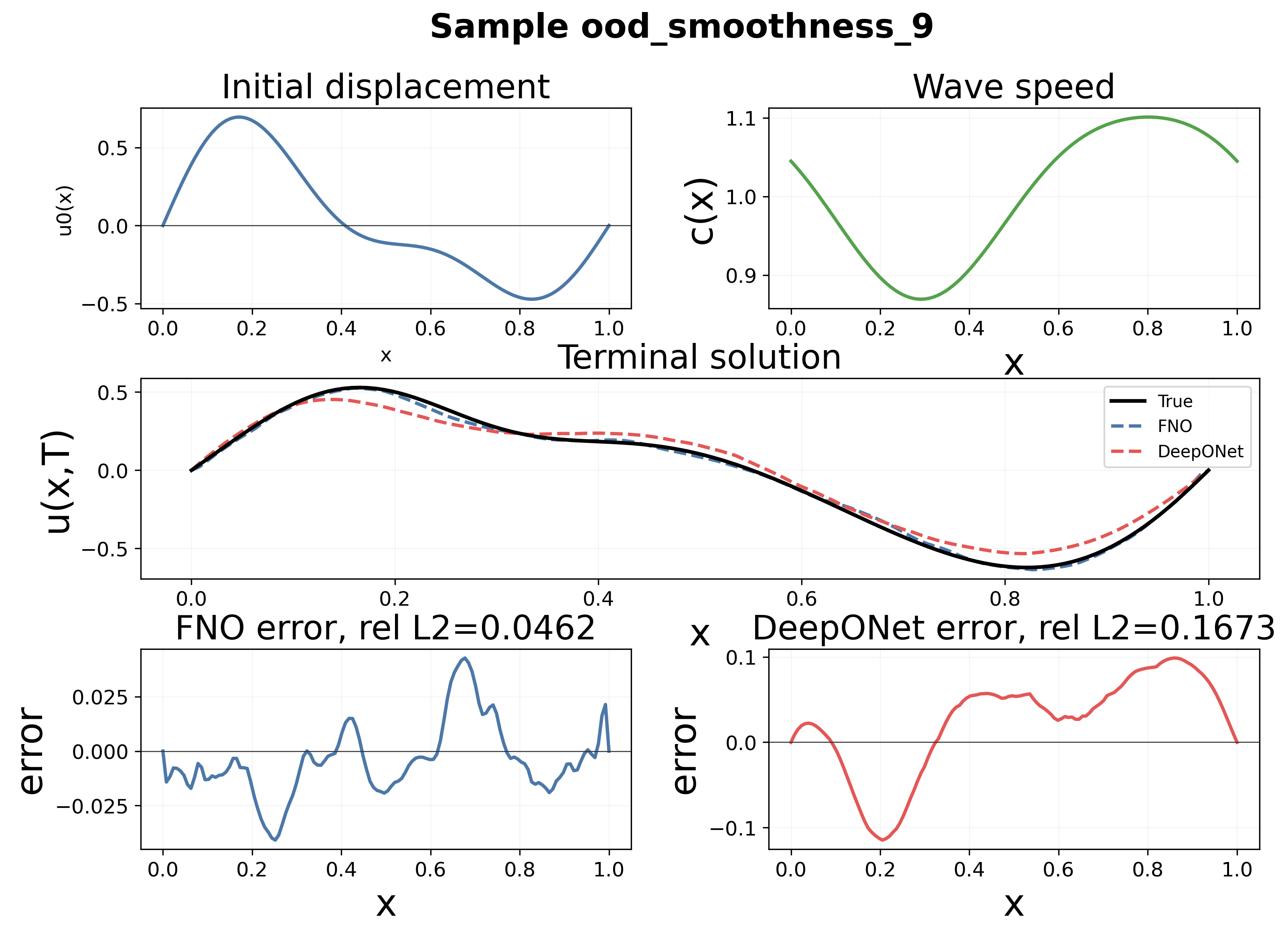}
    \includegraphics[width=0.32\linewidth]{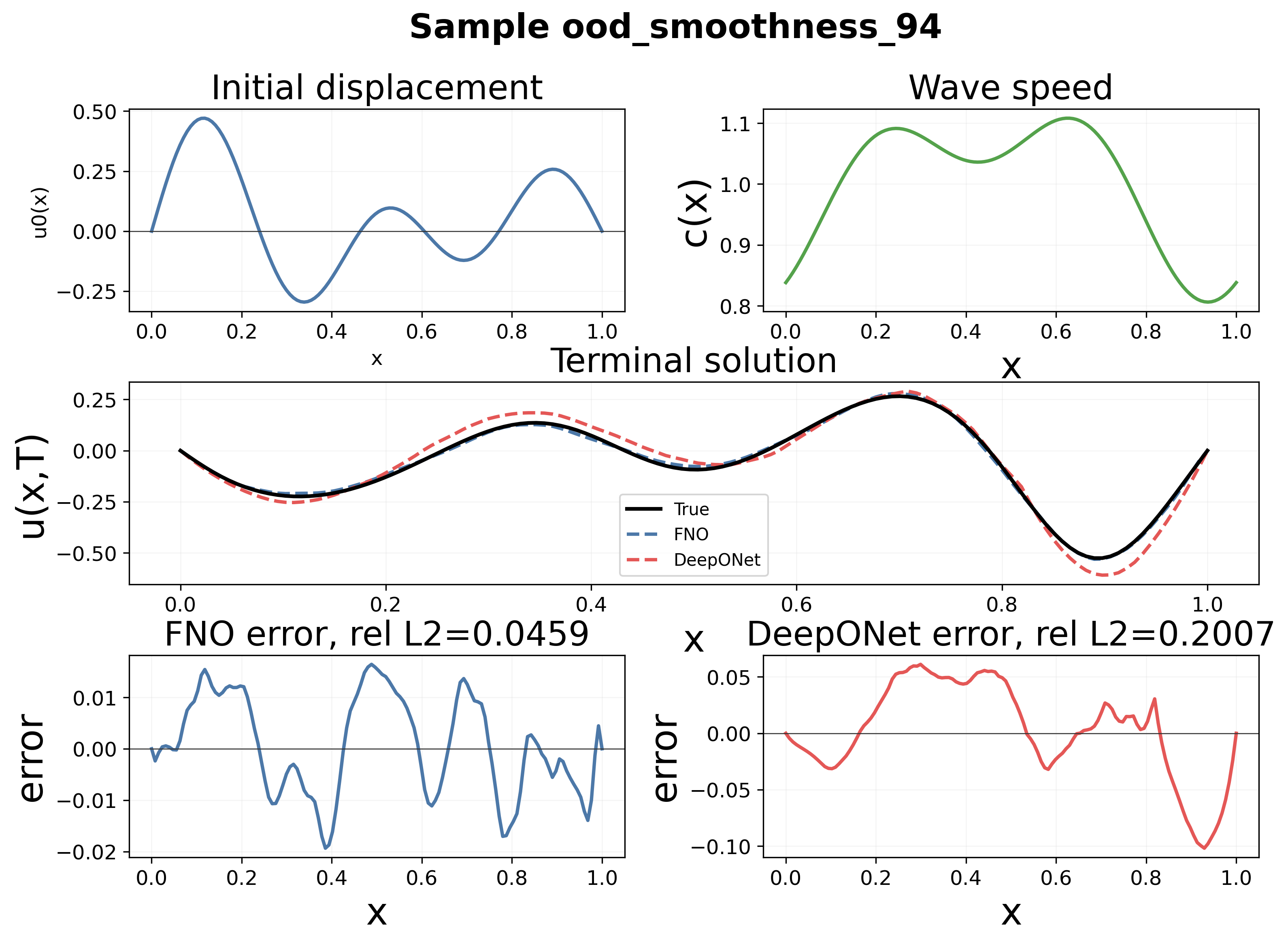}
    \includegraphics[width=0.32\linewidth]{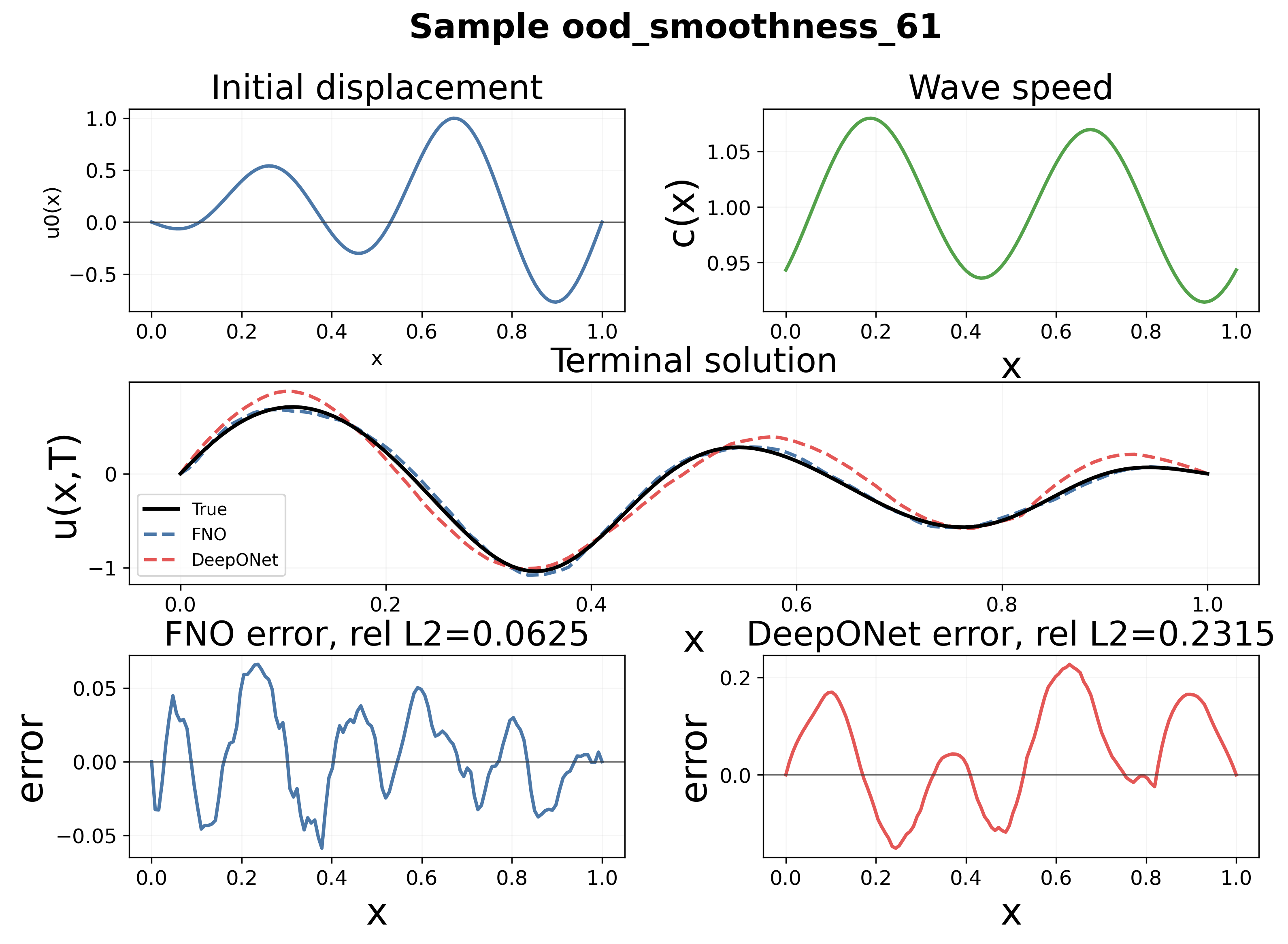}
    \caption{Additional OOD-smoothness prediction examples under shifted coefficient smoothness. Each panel follows the same convention as Figure~\ref{fig:appendix_frequency_ood_examples}, showing input fields, terminal prediction, and residual curves for an independently selected test sample.}
    \label{fig:appendix_smoothness_ood_examples}
\end{figure}

Figure~\ref{fig:appendix_smoothness_ood_examples} shows additional samples from the wave-speed smoothness OOD split. These cases complement the main representative example and show that the lower smoothness-shift error is not restricted to a single selected sample.

\subsection{Full Modes Ablation Table}

Table~\ref{tab:appendix_modes_ablation} reports the complete retained-mode ablation values used to generate the ablation figures in the main text.

\begin{table}[h]
\centering
\caption{Full FNO retained Fourier modes ablation. Values correspond to mean relative \(L_2\) error across each evaluation split.}
\label{tab:appendix_modes_ablation}
\begin{tabular}{ccccc}
\toprule
Retained modes & Validation & ID & OOD-frequency & OOD-smoothness \\
\midrule
8  & 0.173674 & 0.207268 & 1.297833 & 0.059433 \\
16 & 0.169010 & 0.198913 & 1.356445 & 0.058813 \\
32 & 0.208489 & 0.238957 & 1.467290 & 0.060244 \\
\bottomrule
\end{tabular}
\end{table}

\subsection{Additional Spectral Error Curves}

\begin{figure}[h]
    \centering
    \includegraphics[width=0.48\linewidth]{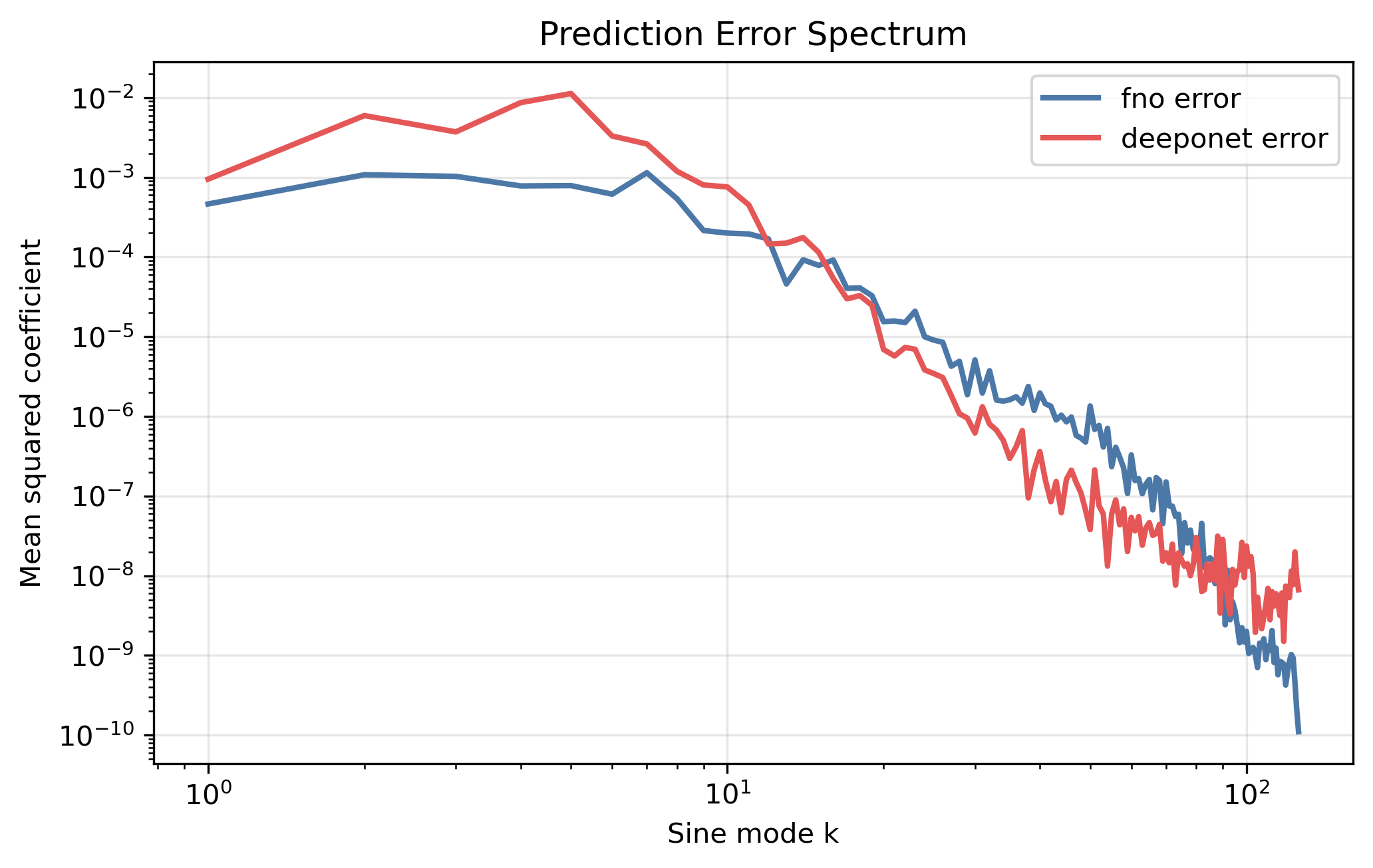}
    \includegraphics[width=0.48\linewidth]{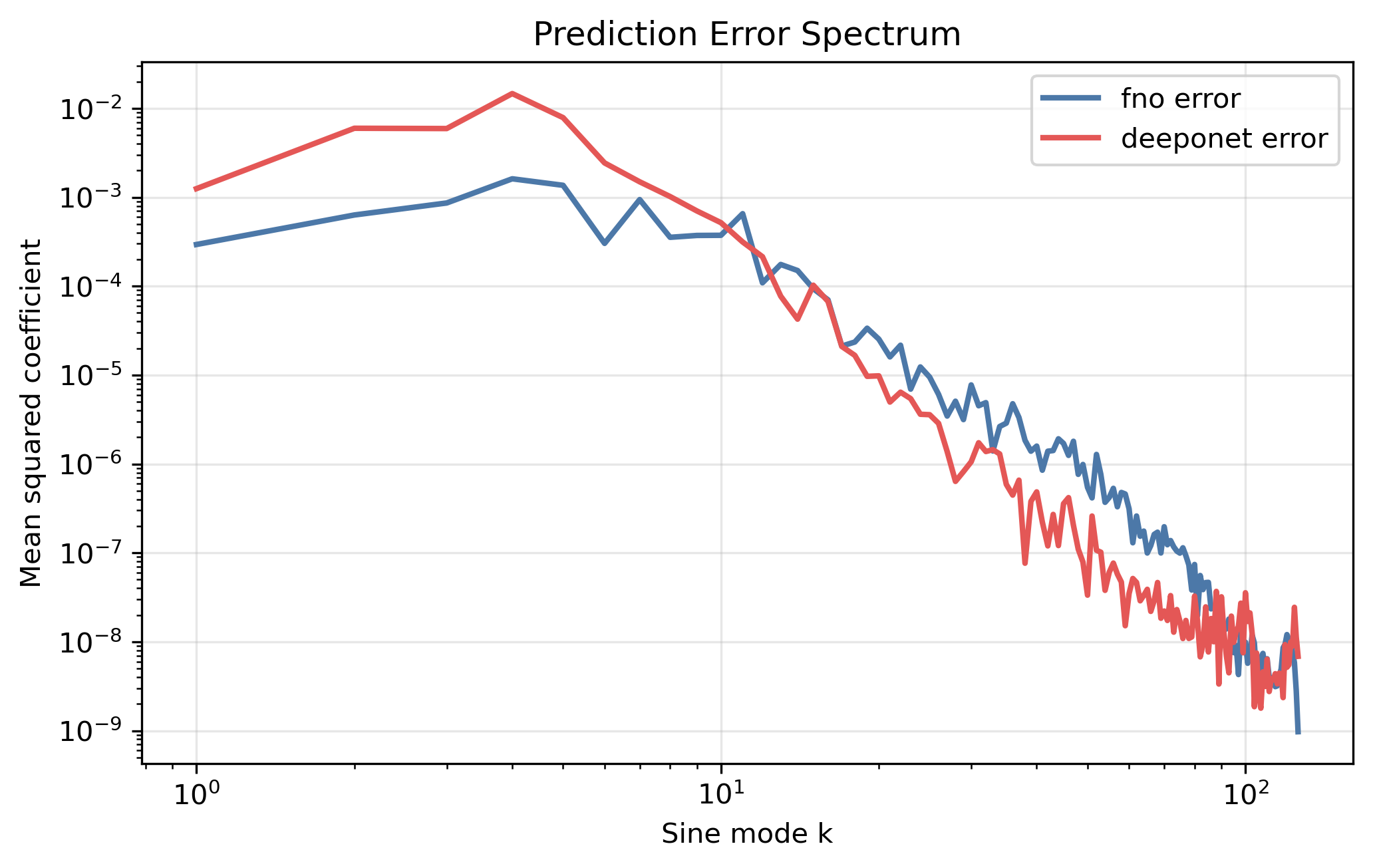}
    \includegraphics[width=0.48\linewidth]{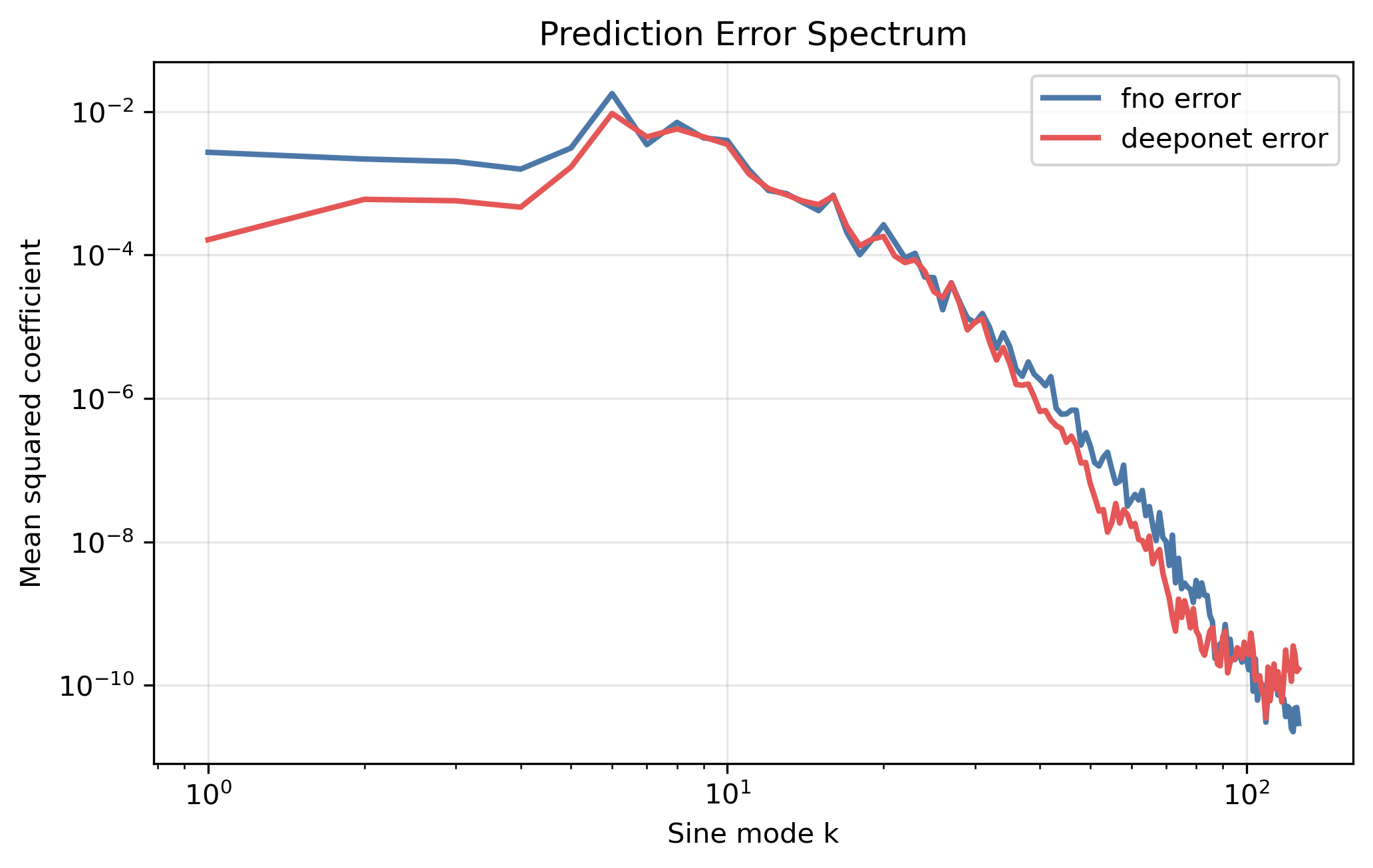}
    \includegraphics[width=0.48\linewidth]{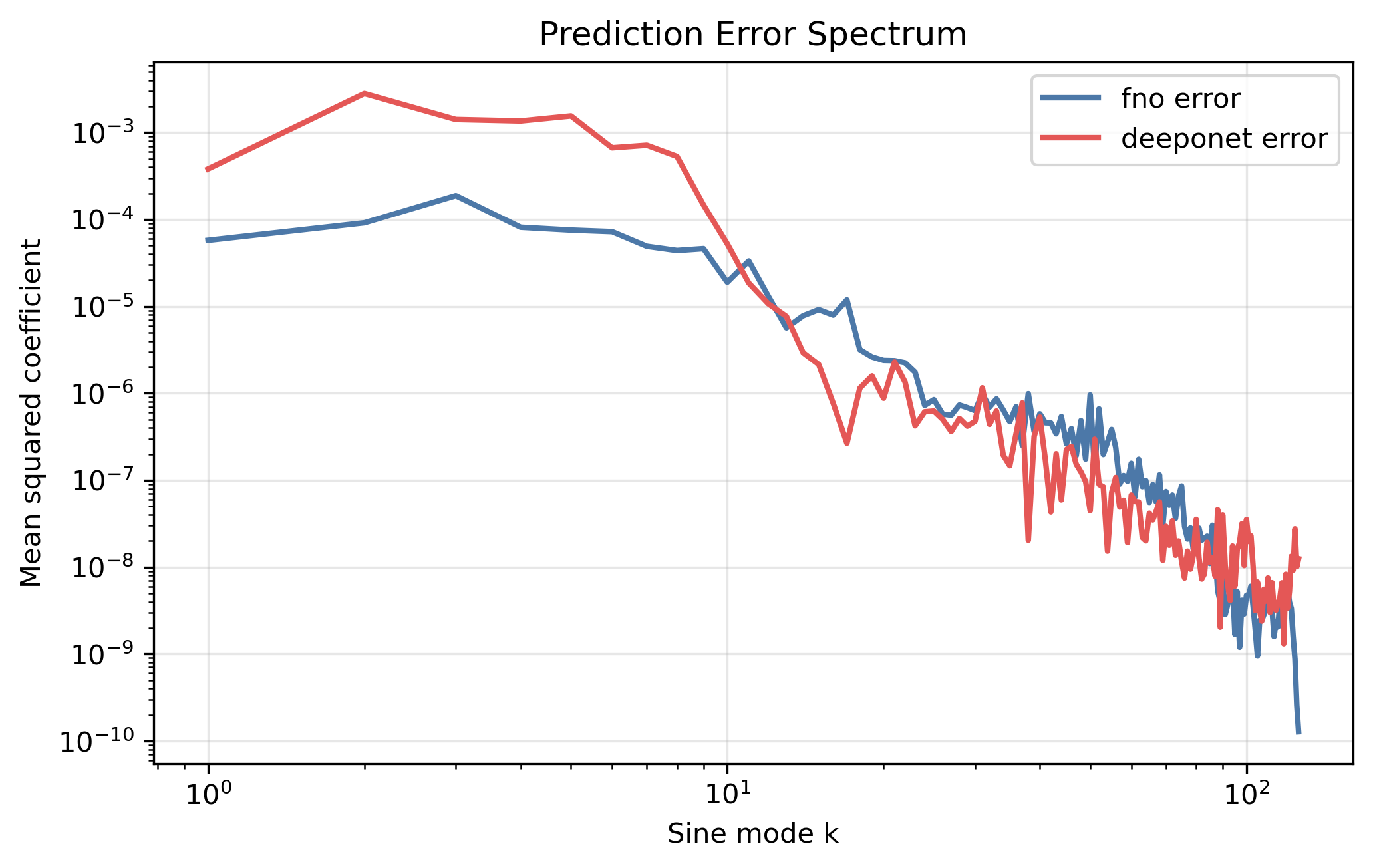}
    \caption{Additional spectral error curves across validation, ID, OOD-frequency, and OOD-smoothness evaluation splits. The curves provide a frequency-domain view of how prediction error varies across sine modes for FNO and DeepONet.}
    \label{fig:appendix_spectral_curves}
\end{figure}

Figure~\ref{fig:appendix_spectral_curves} reports spectral error curves for all main evaluation splits. These plots provide an additional frequency-domain view of the same model behavior reported in the main text.

\end{document}